\documentclass[10pt,twocolumn,letterpaper]{article}

\usepackage{cvpr}
\usepackage{times}
\usepackage{epsfig}
\usepackage{graphicx}
\usepackage{amsmath}
\usepackage{amssymb}
\usepackage{color}
\usepackage{svg}
\usepackage{floatrow}
\usepackage{blindtext}
\usepackage{pstricks-add}
\usepackage{subfig}
\usepackage{multirow}
\usepackage{xcolor,colortbl}
\usepackage{array}
\usepackage{bbm}
\usepackage{mathtools}
\usepackage{enumitem}
\usepackage[pagebackref=true,breaklinks=true,letterpaper=true,colorlinks,bookmarks=false]{hyperref}

\newcolumntype{P}[1]{>{\centering\arraybackslash}p{#1}}
\newcolumntype{M}[1]{>{\centering\arraybackslash}m{#1}}

\definecolor{AMIR}{rgb}{1,0,0}
\definecolor{SASHA}{rgb}{0,0.5,0}
\definecolor{WILL}{rgb}{0,0,1}
\definecolor{red}{rgb}{1,0,0}

\newcommand{\defeq}{\vcentcolon=}

\usepackage[breaklinks=true,bookmarks=false]{hyperref}

\cvprfinalcopy 

\def\cvprPaperID{744} 

\makeatletter
\let\suppsavedmaketitle\maketitle
\let\suppsavedatmaketitle\@maketitle
\let\suppsavedtitle\title
\let\suppsavedauthor\author
\let\suppsavedand\and
\let\suppsavedthanks\thanks
\let\suppsaveddate\date
\makeatother
\DeclareRobustCommand{\suppref}[1]{{\hypersetup{linkcolor=magenta}\hyperlink{supplement.start}{#1}}}

\makeatletter\def\@fnsymbol#1{\ensuremath{\ifcase#1\or *\or \dagger\or \ddagger\or \mathsection\or \mathparagraph\or \|\or **\or \dagger\dagger\or \ddagger\ddagger \else\@ctrerr\fi}}\makeatother
\begin{document}
	
	\title{Taskonomy: Disentangling Task Transfer Learning}
	
	\author{Amir R. Zamir$^{1,2}$ \;\;\hspace{-1mm} Alexander Sax$^{1}$\thanks{Equal.} \;\;\hspace{-1mm} William Shen$^{1*}$ \;\; \hspace{-3mm} Leonidas Guibas$^{1}$ \;\;\hspace{-3mm} Jitendra Malik$^{2}$ \;\; \hspace{-4mm} Silvio Savarese$^{1}$\vspace{10pt}\\ 
		$^1$ Stanford University  \;\;  
		$^2$ University of California, Berkeley\vspace{10pt}\\ 
		\textcolor{blue}{\url{http://taskonomy.stanford.edu/}\vspace{-9pt}}
	}
	
	\maketitle

	\begin{abstract}
		Do visual tasks have a relationship, or are they unrelated? For instance, could having surface normals simplify estimating the depth of an image? Intuition answers these questions positively, implying existence of a \textbf{structure} among visual tasks. Knowing this structure has notable values; it is the concept underlying transfer learning and provides a principled way for identifying redundancies across tasks, in order to, for instance, seamlessly reuse supervision among related tasks or solve many tasks in one system without piling up the complexity.  
		
		We propose a fully computational approach for modeling the structure of the space of visual tasks. This is done via finding (first and higher-order) transfer learning dependencies across a dictionary of twenty-six 2D, 2.5D, 3D, and semantic tasks  in a latent space. The product is a computational taxonomic map for task transfer learning. We study the consequences of this structure, e.g. nontrivial emerged relationships, and exploit them to reduce the demand for labeled data. For example, we show that the total number of labeled datapoints needed for solving a set of 10 tasks can be reduced by roughly $\frac{2}{3}$ (compared to training independently) while keeping the performance nearly the same. 
		We provide a set of tools for computing and probing this taxonomical structure including a solver that users can employ to devise efficient supervision policies for their use cases.   
	\end{abstract}
	
	
	\begin{figure}
		\includegraphics[width=1\columnwidth]{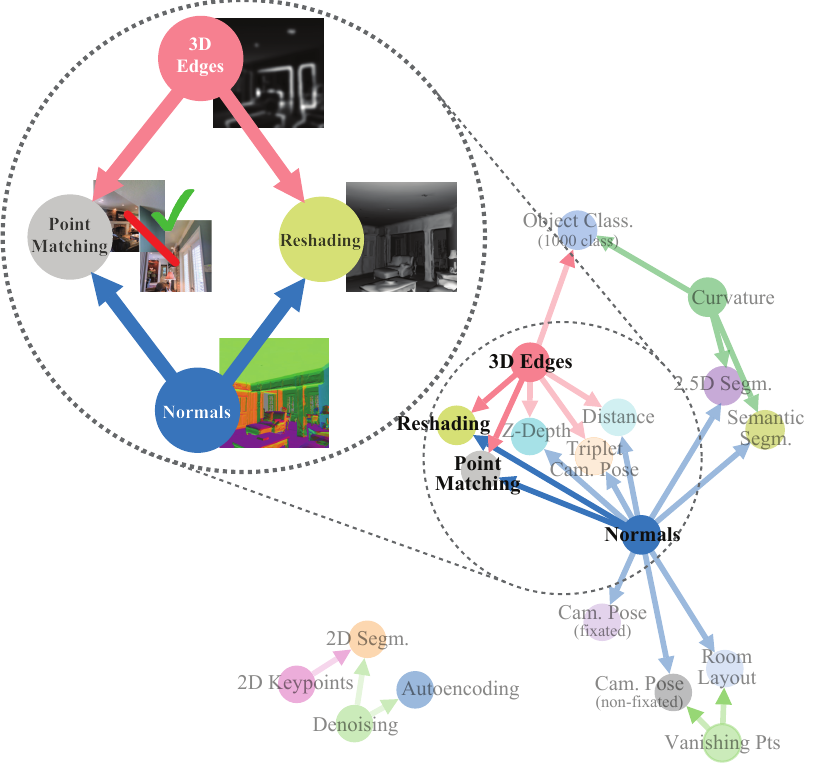}
		\caption{\footnotesize{\textbf{A sample task structure discovered by the computational \textbf{task} tax\textbf{onomy} (\emph{\textbf{taskonomy}})}. It found that, for instance, by combining the learned features of a surface normal estimator and occlusion edge detector, good networks for reshading and point matching can be rapidly trained with little labeled data.}}
		\label{fig:pull}
	\end{figure}
	\section{Introduction}
	Object recognition, depth estimation, edge detection, pose estimation, etc are examples of common vision tasks deemed useful and tackled by the research community. 
	Some of them have rather clear relationships: we understand that surface normals and depth are related (one is a derivate of the other), or vanishing points in a room are useful for orientation and layout estimation. Other relationships are less clear: how edge detection and the shading in a room can, together, perform pose estimation. 
	
	The field of computer vision has indeed gone far without explicitly using these relationships. We have made remarkable progress by developing advanced learning machinery (e.g. ConvNets) capable of finding complex mappings from $X$ to $Y$ when many pairs of $(x,y)$ s.t. $x \in X, y \in Y$ are given as training data. This is usually referred to as fully supervised learning and
	often leads to problems being solved in isolation. Siloing tasks makes training a new task or a comprehensive perception system a Sisyphean challenge, whereby each task needs to be learned individually from scratch. Doing so ignores their quantifiably useful relationships leading to a massive labeled data requirement. 
	
	Alternatively, a model aware of the relationships among tasks demands less supervision, uses less computation, and behaves in more predictable ways. Incorporating such a structure is the first stepping stone towards developing provably efficient comprehensive/universal perception models~\cite{ge2013provable,arora2014provable}, i.e. ones that can solve a large set of tasks before becoming intractable in supervision or computation demands.
	However, this task space structure and its effects are still largely unknown. The relationships are non-trivial, and finding them is complicated by the fact that we have imperfect learning models and optimizers.
	In this paper, we attempt to shed light on this underlying structure and present a framework for mapping the space of visual tasks. Here what we mean by ``structure" is a collection of computationally found relations specifying which tasks supply useful information to another, and by how much (see Fig.~\ref{fig:pull}).

	We employ a fully computational approach for this purpose, with neural networks as the adopted computational function class.  
	In a feedforward network, each layer successively forms more abstract representations of the input containing the information needed for mapping the input to the output. These representations, however, can transmit statistics useful for solving other outputs (tasks), presumably if the tasks are related in some form \cite{sharif2014cnn,donahue2014decaf,LiuLS15Extracted,extractFeatuersRemoteSensing}. This is the basis of our approach: we compute an affinity matrix among tasks based on whether the solution for one task can be sufficiently easily read out of the representation trained for another task. Such transfers are exhaustively sampled, and a Binary Integer Programming formulation extracts a globally efficient transfer policy from them. We show this model leads to solving tasks with far less data than learning them independently and the resulting structure holds on common datasets (ImageNet~\cite{russakovsky2015imagenet} and Places~\cite{zhou2014learning}).

	Being fully computational and representation-based, the proposed approach avoids imposing prior (possibly incorrect) assumptions on the task space. This is crucial because the priors about task relations are often derived from either human intuition or analytical knowledge, while neural networks need not operate on the same principles~\cite{mccloskey:catastrophic, french-1999, Graves2014NeuralTM, Hoshen2015VisualLO, DBLP:journals/corr/ZhangBHRV16, Szegedy2013adversarial}. For instance, although we might expect depth to transfer to surface normals better (derivatives are easy), the opposite is found to be the better direction in a computational framework (i.e. suited neural networks better).

	
	An interactive taxonomy solver which uses our model to suggest data-efficient curricula, a live demo, dataset, and code are available at \href{http://taskonomy.stanford.edu/}{http://taskonomy.stanford.edu/}. 

	\section{Related Work}
	Assertions of existence of a structure among tasks date back to the early years of modern computer science, e.g. with Turing arguing for using learning elements~\cite{turing1950computing,winograd1991thinking} rather than the final outcome or Jean Piaget's works on developmental stages using previously learned stages as sources~\cite{piaget1952origins,gopnik1999scientist, gopnik2004theory}, and have extended to recent works~\cite{richter2017playing,pentina2017multi,kokkinos2016ubernet,doersch2017multi, wang2017transitive, malik2016three, bilen2016integrated, misra2016cross}. Here we make an attempt to actually find this structure. We acknowledge that this is related to a breadth of topics, e.g. compositional modeling~\cite{geman2002composition, bienenstock1997compositionality, boiman2007similarity, faktor2012clustering,lake2016building, tervo2016toward, tenenbaum2011grow}, homomorphic cryptography~\cite{henry2008theory}, lifelong learning~\cite{tessler2017deep, chen2016lml, Silver13lifelongmachine, silver2008guest}, functional maps~\cite{ovsjanikov2012functional}, certain aspects of Bayesian inference and Dirichlet processes~\cite{lake2015human, Tenenbaum06theorybasedbayesian, tenenbaum2011grow, tenenbaum_griffiths_2001, gopnik2004causal, gopnik1999scientist}, few-shot learning~\cite{salakhutdinov2012one,fei2006one, fe2003bayesian,norouzi2013zero, socher2013zero}, transfer learning~\cite{pratt1993discriminability,silver2008guest,finn2016deep, mihalkova2007mapping, niculescu2007inductive, luolabel}, un/semi/self-supervised learning~\cite{erhan2010does, bengio2013representation, doersch2015unsupervised,zhang2016colorful,donahue2014decaf,sharif2014cnn}, which are studied across various fields~\cite{pentina2017multi, thrun2012learning, bingel2017identifying}. We review the topics most pertinent to vision within the constraints of space:

	\textbf{Self-supervised learning} methods leverage the inherent relationships between tasks to learn a desired expensive one (e.g. object detection) via a cheap surrogate (e.g. colorization)~\cite{NorooziF16, pathak2016context, doersch2015unsupervised,zhang2016colorful,amir2016generic,noroozi2017representation}. Specifically, they use a manually-entered local part of the structure in the task space (as the surrogate task is manually defined). In contrast, our approach models this large space of tasks in a computational manner and can discover obscure relationships. 
	
	\textbf{Unsupervised learning} is concerned with the redundancies in the input domain and leveraging them for forming compact representations, which are usually agnostic to the downstream task~\cite{bengio2013representation,kingma2013auto,donahue2016adversarial,berkhin2006survey, fodor2002survey, roweis2000nonlinear}. Our approach is not unsupervised by definition as it is not agnostic to the tasks. Instead, it models the space tasks belong to and in a way utilizes the \emph{functional} redundancies among tasks.

	\textbf{Meta-learning} generally seeks performing the learning at a level higher than where conventional learning occurs, e.g. as employed in reinforcement learning~\cite{duan2016rl, finn2017oneshot, finn2016irl}, optimization~\cite{andrychowicz2016learning, schulman2015trpo, kingma2014adam}, or certain architectural mechanisms~\cite{finn2017model, finn16semisuprl, srivastava2014dropout, mikolov2013languagepairs}. The motivation behind meta learning has similarities to ours and our outcome can be seen as a computational meta-structure of the space of tasks. 
	
	\begin{figure*}
		\includegraphics[width=1\textwidth]{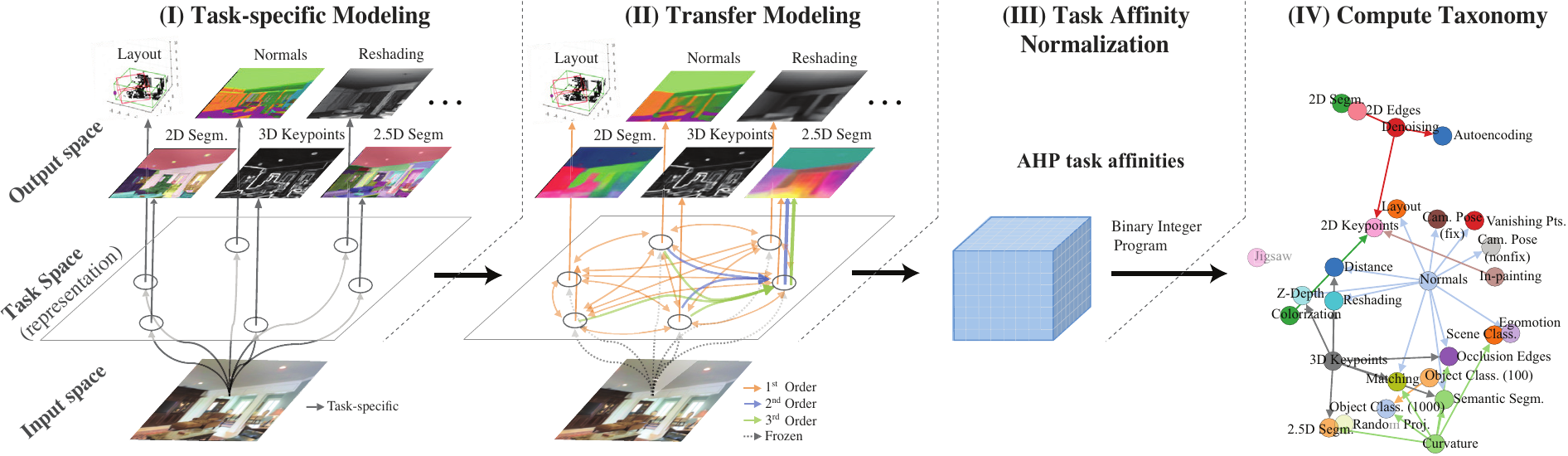}
		\caption{\footnotesize{\textbf{Computational modeling of task relations and creating the taxonomy.} From left to right: I. Train task-specific networks. II. Train (first order and higher) transfer functions among tasks in a latent space. III. Get normalized transfer affinities using AHP (Analytic Hierarchy Process). IV. Find global transfer taxonomy using BIP (Binary Integer Program).}\vspace{-6pt}}
		 \vspace{-6pt}
		\label{fig:process}
	\end{figure*}

	\textbf{Multi-task learning} targets developing systems that can provide multiple outputs for an input in one run~\cite{kokkinos2016ubernet, doersch2017multi}. Multi-task learning has experienced recent progress and the reported advantages are another support for existence of a useful structure among tasks~\cite{tessler2017deep, amir2016generic, kokkinos2016ubernet, richter2017playing, pentina2017multi,kokkinos2016ubernet,doersch2017multi, wang2017transitive, malik2016three, bilen2016integrated, misra2016cross}. Unlike multi-task learning, we explicitly model the relations among tasks and extract a meta-structure. The large number of tasks we consider also makes developing one multi-task network for all infeasible.
	
	%

	\textbf{Domain adaption} seeks to render a function that is developed on a certain domain applicable to another~\cite{hoffman2014continuous,yang2007adapting, aytar2011tabula, saenko2010adapting, kulis2011you, fernando2013unsupervised, gopalan2011domain}. It often addresses a shift in the \emph{input} domain, e.g. webcam images to D-SLR~\cite{jhuo2012robust}, while the task is kept the same. In contrast, our framework is concerned with \emph{output} (task) space, hence can be viewed as \emph{task/output adaptation}. We also perform the adaptation in a larger space among many elements, rather than two or a few.
	
	In the context of our approach to modeling transfer learning across tasks:
	
	\textbf{Learning Theoretic} approaches may overlap with any of the above topics and usually focus on providing generalization guarantees. They vary in their approach: e.g. by modeling transferability with the transfer family required to map a hypothesis for one task onto a hypothesis for another~\cite{Ben-David2008}, through information-based approaches~\cite{Mahmud2007}, or through modeling inductive bias~\cite{Baxter:1997:BTM:262868.262870}. For these guarantees, learning theoretic approaches usually rely on intractable computations, or avoid such computations by restricting the model or task. Our method draws inspiration from theoretical approaches but eschews (for now) theoretical guarantees in order to use modern neural machinery.

	\section{Method}
	\label{sec:method}


	We define the problem as follows: we want to maximize the collective performance on a set of tasks $\mathcal{T} =  \{t_1,...,t_n\}$, subject to the constraint that we have a limited supervision budget $\gamma$ (due to financial, computational, or time constraints). 
	We define our supervision budget $\gamma$ to be the maximum allowable number of tasks that we are willing to train from scratch (i.e. \emph{source} tasks).
	The task dictionary is defined as $\mathcal{V}$=$\mathcal{T} \cup \mathcal{S}$ where $\mathcal{T}$ is the set of tasks which we want solved (\emph{target}), and $\mathcal{S}$ is the set of tasks that can be trained (\emph{source}). Therefore, $\mathcal{T} - \mathcal{T}\cap\mathcal{S}$ are the tasks that we want solved but cannot train (``target-only"), $\mathcal{T}\cap\mathcal{S}$ are the tasks that we want solved but could play as source too, and $\mathcal{S} - \mathcal{T}\cap\mathcal{S}$ are the ``source-only" tasks which we may not directly care about to solve (e.g. jigsaw puzzle) but can be optionally used if they increase the performance on~$\mathcal{T}$.

	The \textbf{task} tax\textbf{onomy} (\emph{\textbf{taskonomy}}) is a computationally found directed hypergraph that captures the notion of task transferability over any given task dictionary. An edge between a group of source tasks and a target task represents a feasible transfer case and its weight is the prediction of its performance. We use these edges to estimate the globally optimal transfer policy to solve $\mathcal{T}$. Taxonomy produces a family of such graphs, parameterized by the available supervision budget, chosen tasks, transfer orders, and transfer functions' expressiveness.

	\begin{figure}
		\includegraphics[width=1\columnwidth]{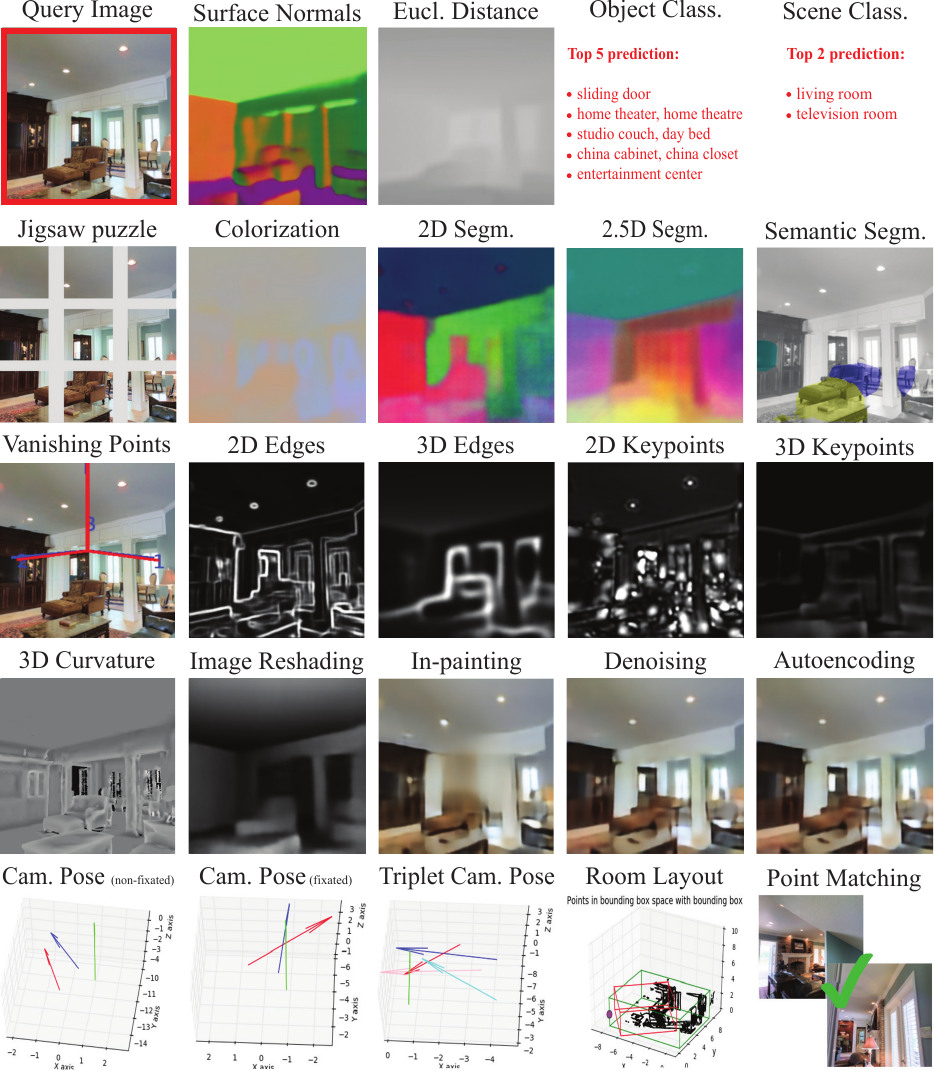}
		\caption{\footnotesize{\textbf{Task Dictionary.} Outputs of 24 (of 26) task-specific networks for a query (top left). See results of applying  frame-wise on a video \href{http://taskonomy.stanford.edu/\#models}{here}.} }
		\label{fig:outputs}
	\end{figure}
	
	Taxonomy is built using a four step process depicted in Fig.~\ref{fig:process}. 
	In stage I, a task-specific network for each task in $\mathcal{S}$ is trained. In stage II, all feasible transfers between sources and targets are trained. We include higher-order transfers which use multiple inputs task to transfer to one target. In stage III, the task affinities acquired from transfer function performances are normalized, and in stage IV, we synthesize a hypergraph which can predict the performance of any transfer policy and optimize for the optimal one.

	A vision task is an abstraction read from a raw image. We denote a task $t$ more formally as a function $f_t$ which maps image $I$ to $f_t(I)$. Our dataset, $\mathcal{D}$, contains for each task $t$ a set of training pairs $(I, f_t(I))$, e.g. $(image,depth)$. 

	
	\textbf{Task Dictionary:} Our mapping of task space is done via (26) tasks included in the dictionary, so we ensure they cover common themes in computer vision (2D, 3D, semantics, etc) to elucidate fine-grained structures of task space. See Fig.~\ref{fig:outputs} for some of the tasks with detailed definition of each task provided in the \suppref{supplementary material}. We include tasks with various levels of abstraction, ranging from solvable by a simple kernel convolved over the image (e.g. edge detection) to tasks requiring basic understanding of scene geometry (e.g. vanishing points) and more abstract ones involving semantics (e.g. scene classification). 
	
	It is critical to note the task dictionary is meant to be a \emph{sampled set, not an exhaustive list}, from a denser space of all conceivable visual tasks/abstractions. Sampling gives us a tractable way to sparsely model a dense space, and the hypothesis is that (subject to a proper sampling) the derived model should generalize to out-of-dictionary tasks. The more regular / better sampled the space, the better the generalization. We evaluate this in Sec.~\ref{sec:novel_task} with supportive results. For evaluation of the robustness of results w.r.t the choice of dictionary, see the \suppref{supplementary material}.
	
	\begin{figure}
		\includegraphics[width=1\columnwidth]{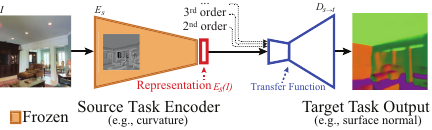}
		\caption{ \footnotesize{\textbf{Transfer Function}. A small readout function is trained to map representations of source task's frozen encoder to target task's labels. If order$>1$, transfer function receives representations from multiple sources.}}
		\label{fig:training_procedure}
	\end{figure}

	\textbf{Dataset:} We need a dataset that has annotations for \emph{every task on every image}. Training all of our tasks on exactly the same pixels eliminates the possibility that the observed transferabilities are affected by different input data peculiarities rather than only task intrinsics. There has not been such a dataset of scale made of real images, 
	so we created a dataset of 4 million images of indoor scenes from about 600 buildings; every image has an annotation for every task. The images are registered on and aligned with building-wide meshes similar to~\cite{armeni2017joint,gibson,chang2017matterport3d} enabling us to programmatically compute the ground truth for many tasks without human labeling. For the tasks that still require labels (e.g. scene classes), we generate them using Knowledge Distillation~\cite{hinton2015distilling} from known methods~\cite{zhou2014learning,lin2014microsoft,li2016fully,russakovsky2015imagenet}. See the \suppref{supplementary material} for full details of the process and a user study on the final quality of labels generated using Knowledge Distillation (showing $<7\%$ error). 
	

	\subsection{Step I: Task-Specific Modeling} 
	We train a fully supervised task-specific network for each task in $\mathcal{S}$. Task-specific networks have an encoder-decoder architecture homogeneous across all tasks, where the encoder is large enough to extract powerful representations, and the decoder is large enough to achieve a good performance but is much smaller than the encoder.

	\subsection{Step II: Transfer Modeling} 
\begin{figure}
		\includegraphics[width=1\columnwidth]{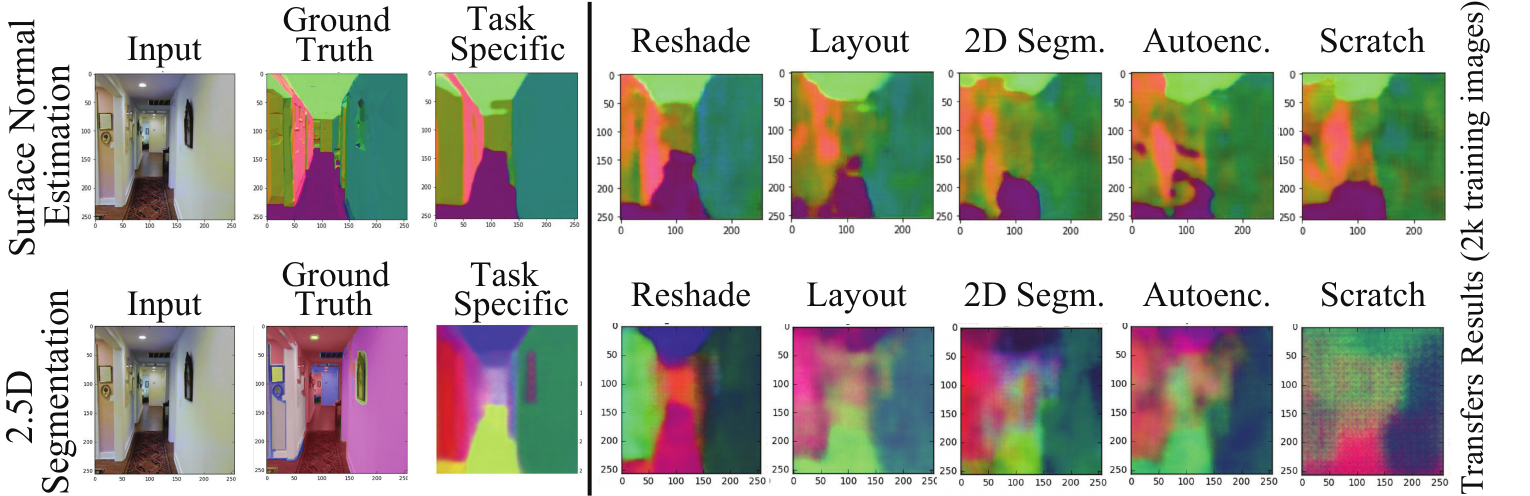}
		\vspace{-0.3cm}		    
		\caption{\footnotesize{\textbf{Transfer results} to normals (upper) and 2.5D Segmentation (lower) from 5 different source tasks. The spread in transferability among different sources is apparent, with reshading among top-performing ones in this case. Task-specific networks were trained on 60x more data. ``Scratch'' was trained from scratch without transfer learning.}}
		\label{fig:sample_transfers}
	\end{figure}

	Given a source task $s$ and a target task $t$, where $s \in \mathcal{S}$ and $t \in \mathcal{T}$, a transfer network learns a small readout function for $t$ given a statistic computed for $s$ (see Fig~\ref{fig:training_procedure}). The statistic is the representation for image $I$ from the encoder of  $s$: $E_s(I)$. The readout function ($D_{s\rightarrow t}$) is parameterized by $\theta_{s \rightarrow t}$ minimizing the loss $L_t$:
	\begin{equation}
	D_{s \rightarrow t} \defeq \arg \min_{\theta} \mathbb{E}_{I \in \mathcal{D}}\Big[L_t\Big(D_{\theta}\big(E_s(I)\big), f_t(I)\Big)\Big],
	\end{equation}
	where $f_t(I)$ is ground truth of $t$ for image $I$. $E_s(I)$ may or may not be sufficient for solving $t$ depending on the relation between $t$ and $s$ (examples in Fig.~\ref{fig:sample_transfers}). Thus, the performance of $D_{s\rightarrow t}$ is a useful metric as task affinity.
	We train transfer functions for all feasible source-target combinations.
	
	\textbf{Accessibility}: For a transfer to be successful, the latent representation of the source should both be \emph{inclusive} of sufficient information for solving the target and have the information \emph{accessible}, i.e. easily extractable (otherwise, the raw image or its compression based representations would be optimal). Thus, it is crucial for us to adopt a low-capacity (small) architecture as transfer function trained with a small amount of data, in order to measure transferability conditioned on being highly accessible. We use a shallow fully convolutional network and train it with little data (8x to 120x less than task-specific networks).
	
\begin{figure}
		\includegraphics[width=1\columnwidth]{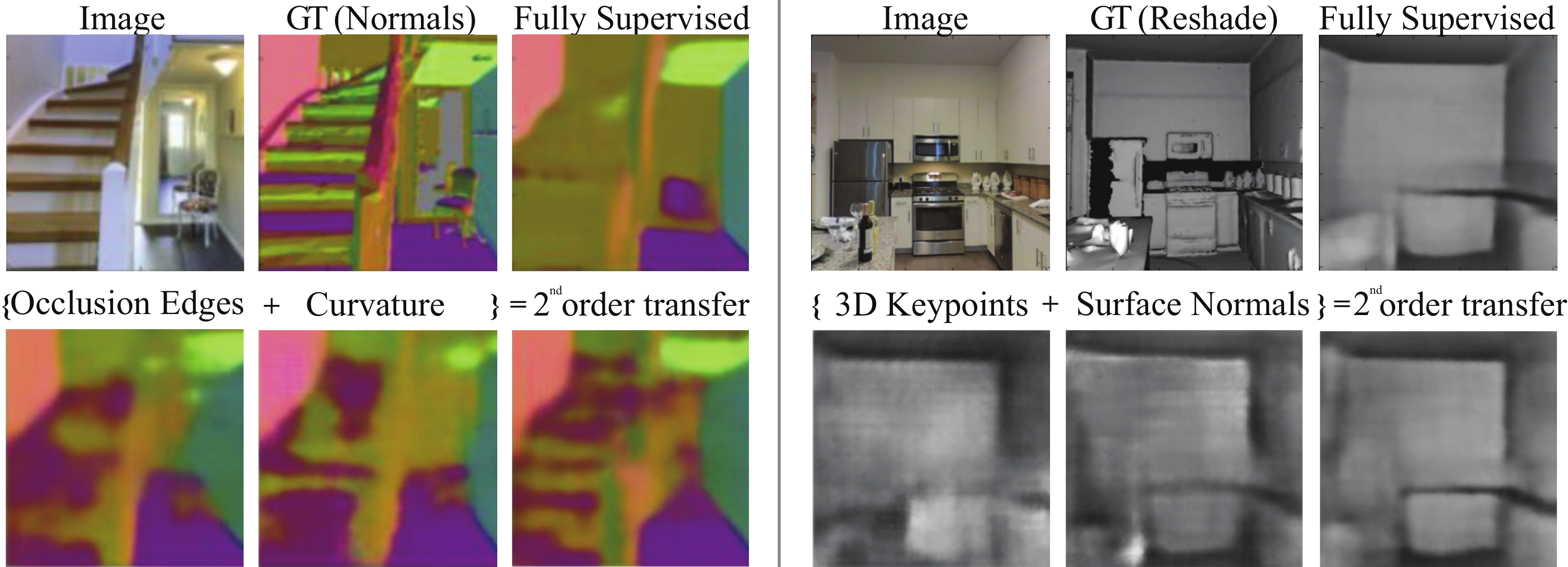}
		\caption{ \footnotesize{\textbf{Higher-Order Transfers.} Representations can contain complementary information. E.g. by transferring simultaneously from 3D Edges and Curvature individual stairs were brought out. See our publicly available interactive \href{http://taskonomy.stanford.edu/transfers}{transfer visualization page} for more examples.}}
		\label{fig:second_order_taxonomy}
	\end{figure}

	\textbf{Higher-Order Transfers:} 
	Multiple source tasks can contain complementary information for solving a target task (see examples in Fig~\ref{fig:second_order_taxonomy}). We include higher-order transfers which are the same as first order but receive multiple representations in the input. Thus, our transfers are functions $D: \wp(\mathcal{S}) \rightarrow \mathcal{T}$, where $\wp$ is the powerset operator.
	
	As there is a combinatorial explosion in the number of feasible higher-order transfers ($|\mathcal{T}| \times {|\mathcal{S}|\choose k}$ for $k^{th}$ order), we employ a sampling procedure with the goal of filtering out higher-order transfers that are less likely to yield good results, without training them. We use a beam search: for transfers of order $k \leq 5$ to a target, we select its 5 best sources (according to $1^{st}$ order performances) and include all of their order-$k$ combination. For $k \geq 5$, we use a beam of size 1 and compute the transfer from the top $k$ sources.

	\textbf{Transitive Transfers:} We examined if transitive task transfers ($s\rightarrow t_1\rightarrow t_2$) could improve the performance over their direct counterpart ($s \rightarrow t_2$), but found that the two had equal performance in almost all cases in both high-data and low-data scenarios. The experiment is provided in the \suppref{supplementary material}. Therefore, we need not consider the cases where branching would be more than one level deep when searching for the optimal transfer path.

	\subsection{Step III: Ordinal Normalization using Analytic Hierarchy Process (AHP)}
\begin{figure}
		\centering
		\includegraphics[width=1\columnwidth]{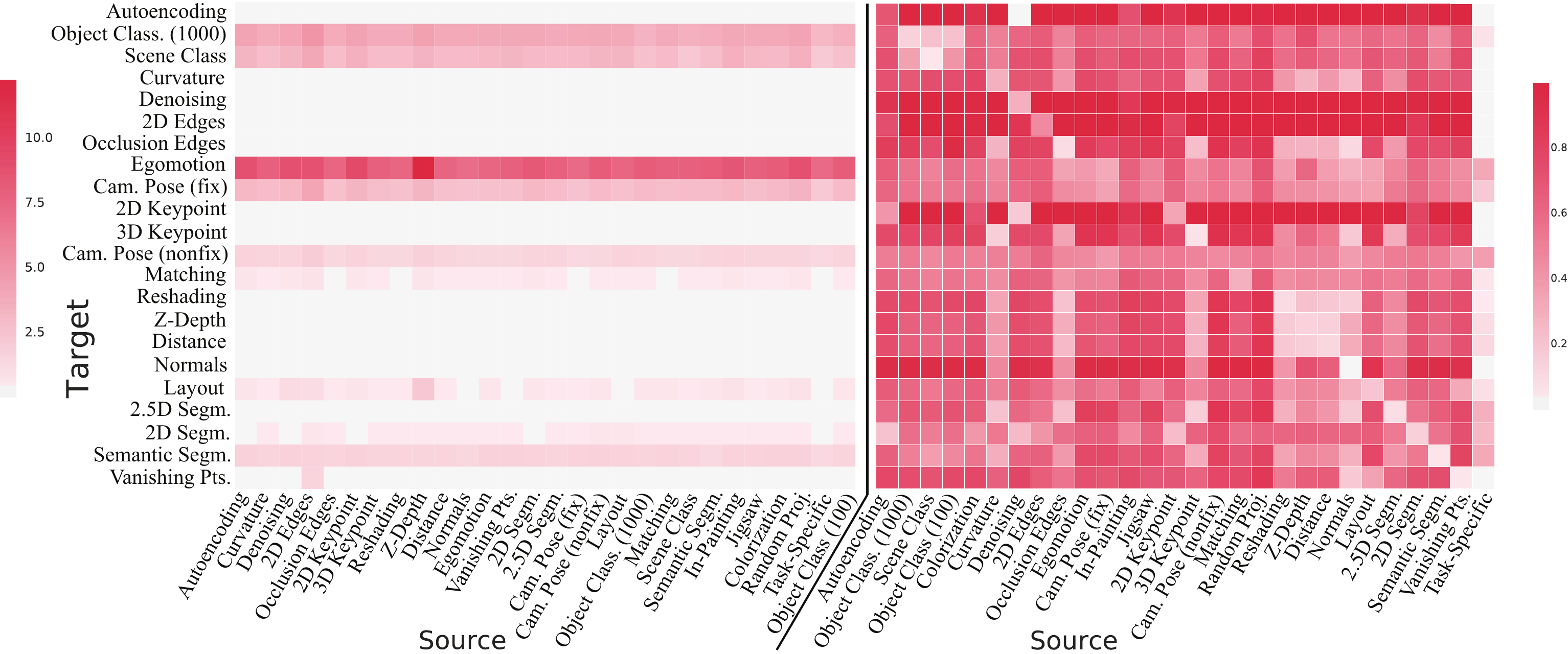}
		\caption{\footnotesize{\textbf{First-order task affinity matrix} before (left) and after (right) Analytic Hierarchy Process (AHP) normalization. Lower means better transfered. For visualization, we use standard affinity-distance method $\textit{dist} = e^{-\beta \cdot P}$ (where $\beta = 20$ and $e$ is element-wise matrix exponential). See \suppref{supplementary material}} for the full matrix with higher-order transfers.}
		\label{fig:task_distances}
	\end{figure}

	We want to have an affinity matrix of transferabilities across tasks. Aggregating the raw losses/evaluations $L_{s\rightarrow t}$ from transfer functions into a matrix is obviously problematic as they have vastly different scales and live in different spaces (see Fig.~\ref{fig:task_distances}-left). 
	Hence, a proper normalization is needed. A naive solution would be to linearly rescale each row of the matrix to the range $[0,1]$. This approach fails when the actual output quality increases at different speeds w.r.t. the loss. As the loss-quality curve is generally unknown, such approaches to normalization are ineffective. 
	
	Instead, we use an \emph{ordinal} approach in which the output quality and loss are only assumed to change monotonically.
	For each $t$, we construct $W_{t}$ a pairwise tournament matrix between all feasible sources for transferring to $t$. 
	The element at $(i, j)$ is the percentage of images in a held-out test set, $\mathcal{D}_{test}$, on which $s_i$ transfered to $t$ better than $s_j$ did (i.e. $D_{s_i \rightarrow t}(I) > D_{s_j \rightarrow t}(I)$).
	
	We clip this intermediate pairwise matrix $W_t$ to be in $[0.001, 0.999]$ as a form of Laplace smoothing. Then we divide $W'_t = W_t / W_t^T$ so that the matrix shows how many times better $s_i$ is compared to $s_j$. The final tournament ratio matrix is positive reciprocal with each element $w'_{i,j}$ of $W'_t$:

	\begin{equation}
	w'_{i,j} = \frac{
		\mathop{\mathbb{E}}_{I \in \mathcal{D_\textit{test}}}[D_{s_i \rightarrow t}(I) > D_{s_j \rightarrow t}(I)]
	}{
		\mathop{\mathbb{E}}_{I \in \mathcal{D_\textit{test}}}[D_{s_i \rightarrow t}(I) < D_{s_j \rightarrow t}(I)]
	}.
	\end{equation}
	
	We quantify the final transferability of $s_i$ to $t$ as the corresponding ($i^{th}$) component of the principal eigenvector of $W'_t$ (normalized to sum to 1). 
	The elements of the principal eigenvector are a measure of centrality, and are proportional to the amount of time that an infinite-length random walk on $W'_t$ will spend at any given source~\cite{randomwalks}. 
	We stack the principal eigenvectors of $W'_t$ for all $t \in \mathcal{T}$, to get an affinity matrix $P$ (`p' for performance)|see Fig.~\ref{fig:task_distances}, right.

	This approach is derived from Analytic Hierarchy Process \cite{AHP}, a method widely used in operations research to create a total order based on multiple pairwise comparisons. 
	

	
	\subsection{Step IV: Computing the Global Taxonomy}\label{test_time}
	\label{sec:bip}
	\setlength\abovedisplayskip{0pt}
	\setlength\belowdisplayskip{3pt}

	Given the normalized task affinity matrix, we need to devise a global transfer policy which maximizes collective performance across all tasks, while minimizing the used supervision. This problem can be formulated as subgraph selection where tasks are nodes and transfers are edges. The optimal subgraph picks the ideal source nodes and the best edges from these sources to targets while satisfying that the number of source nodes does not exceed the supervision budget. We solve this subgraph selection problem using Boolean Integer Programming (BIP), described below, which can be solved optimally and efficiently~\cite{gurobi,cplex2009v12}.
	
	Our transfers (edges), $E$, are indexed by $i$ with the form $(\{s^i_1, \dotso, s^i_{m_i}\}, t^i)$ where $\{s^i_1, \dotso ,s^i_{m_i}\} \subset \mathcal{S}$ and $t^i \in \mathcal{T}$. We define operators returning target and sources of an edge:
	
	\begin{align*}
	\big(\{s^i_1, \dotso, s^i_{m_i}\}, t^i\big) \hspace{2mm} & \xmapsto[]{sources} \hspace{1mm} \{s^i_1, \dotso, s^i_{m_i}\}\\
	\big(\{s^i_1, \dotso, s^i_{m_i}\}, t^i\big) \hspace{2mm} & \xmapsto[]{target} \hspace{4mm} t^i.
	\end{align*}
	Solving a task $t$ by fully supervising it is denoted as $\big(\{t\}, t\big)$. We also index the targets $\mathcal{T}$ with $j$ so that in this section, $i$ represents edges and $j$ represents targets. 
	
	The parameters of the problem are:
	the supervision budget ($\gamma$) and a measure of performance on a target from each of its transfers ($p_i$), i.e. the affinities from $P$. We can also optionally include additional parameters of: $r_j$ specifying the relative importance of each target task and $\ell_i$ specifying the relative cost of acquiring \textit{labels} for each task. 
	
	The BIP is parameterized by a vector $x$ where each transfer and each task is represented by a binary variable; $x$ indicates which nodes are picked to be source and which transfers are selected. The canonical form for a BIP is: 
	\begin{align*}
	\text{maximize: } & c^T x\,,\\
	\text{subject to: } & Ax \preceq b 
	\text{ and } x \in \{0, 1\}^{|E| + |\mathcal{V}|}\,.
	\end{align*}
	
	Each element $c_i$ for a transfer is the product of the importance of its target task and its transfer performance:
	\begin{equation}
	c_i := r_{\textit{target}(i)} \cdot p_i\,.
	\end{equation}
	Hence, the \emph{collective} performance on all targets is the summation of their individual AHP performance, $p_i$, weighted by the user specified importance, $r_i$.
	
	Now we add three types of constraints via matrix $A$ to enforce each feasible solution of the BIP instance corresponds to a valid subgraph for our transfer learning problem: \emph{Constraint I:} if a transfer is included in the subgraph, all of its source nodes/tasks must be included too, \emph{Constraint II:}  each target task has exactly one transfer in, \emph{Constraint III:} supervision budget is not exceeded. 
	
	\emph{Constraint I}: For each row $a_i$ in $A$ we require $a_i \cdot x \leq b_i$, where 
	\begin{align}
	a_{i,k} &\defeq \begin{cases}
	|\textit{sources}(i)| & \text{if } k = i\\
	-1 & \text{if } (k - |E|) \in \textit{sources}(i)\\
	0 & \text{otherwise}\\
	\end{cases}\\
	b_{i} &= 0.
	\end{align}
	
	\emph{Constraint II}: Via two rows $a_{|E| + j}$ and $a_{|E| + j + 1}$, we enforce that target $j$ has exactly one transfer:
	\begin{equation}
    \begin{array}{l}
	    a_{|E| + j, i} \hspace{10.2pt} \defeq \hspace{7.3pt} \mathbbm{1}_{\{\textit{target}(i)=j\}},\hspace{0.5cm} b_{|E| + j} \hspace{10.2pt} \defeq +1, \\ 
	    a_{|E| + j+1, i} \defeq -\mathbbm{1}_{\{\textit{target}(i)=j\}},\hspace{0.5cm} b_{|E| + j+1} \defeq -1.
    \end{array}
	\end{equation}

	\emph{Constraint III}: the solution is enforced to not exceed the budget. Each transfer $i$ is assigned a label cost $\ell_i$, so 
	\begin{equation}
	a_{|E| + 2|\mathcal{V}| + 1, i}\defeq \ell_i,\hspace{0.5cm} b_{|E| + 2|\mathcal{V}| + 1} \defeq \gamma.
	\end{equation}
	
	The elements of A not defined above are set to 0. The problem is now a valid BIP and can be optimally solved in a fraction of a second~\cite{gurobi}. 
	The BIP solution $\hat{x}$ corresponds to the optimal subgraph, which is our taxonomy.

	\begin{figure*}
		\includegraphics[width=0.95\textwidth]{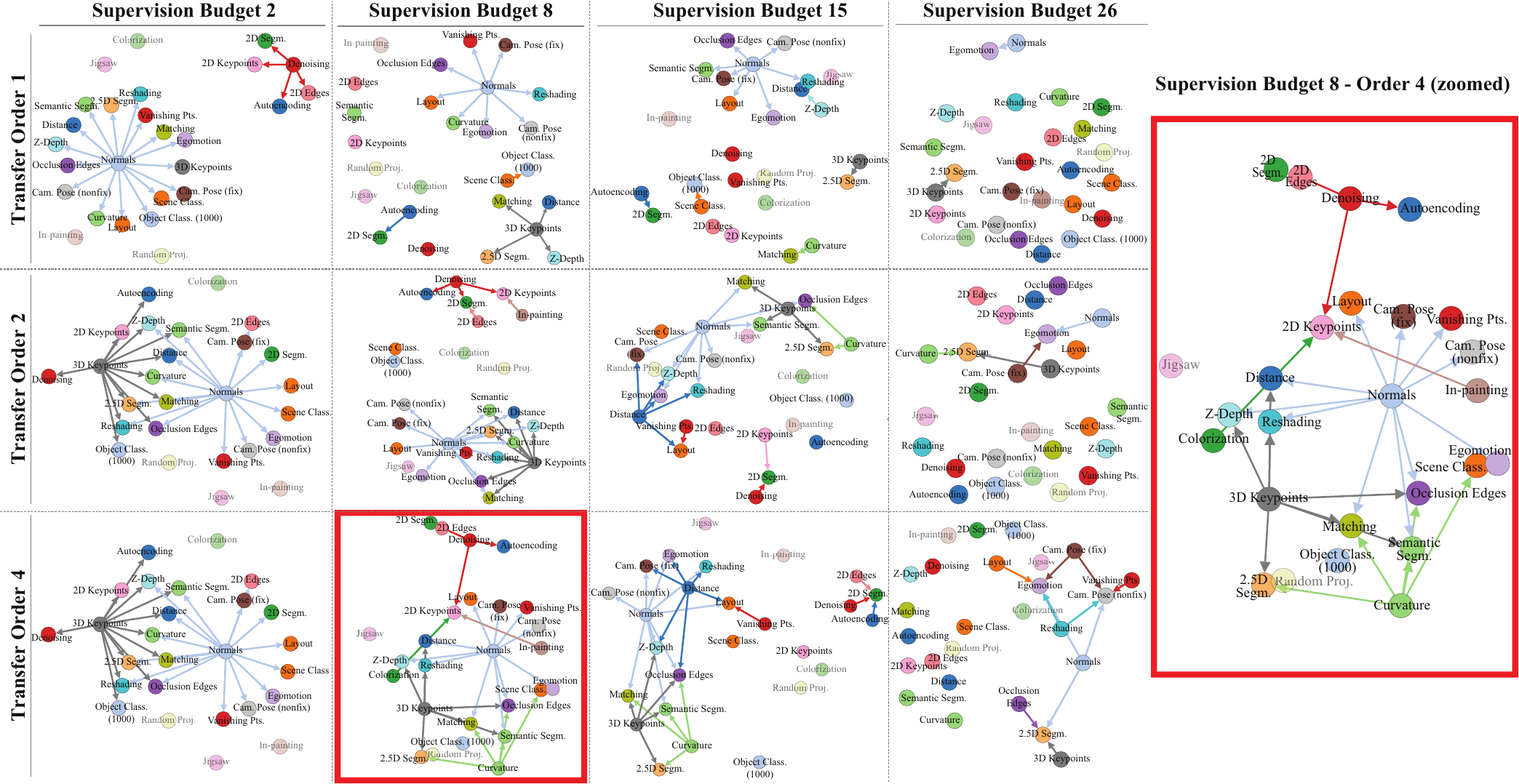}
		\caption{\footnotesize{\textbf{Computed taxonomies} for solving 22 tasks given various supervision budgets (x-axes), and maximum allowed transfer orders (y-axes). One is magnified for better visibility. Nodes with incoming edges are target tasks, and the number of their incoming edges is the order of their chosen transfer function. Still transferring to some targets when tge budget is 26 (full budget) means certain transfers started performing better than their fully supervised task-specific counterpart. See the interactive \href{http://taskonomy.stanford.edu/api}{solver website} for color coding of the nodes by \emph{Gain} and \emph{Quality} metrics. Dimmed nodes are the source-only tasks, and thus, only participate in the taxonomy if found worthwhile by the BIP optimization to be one of the sources.}}
    	\vspace{-0.3cm}
		\label{fig:taxonomies_over_time}
	\end{figure*}
	
	\section{Experiments}
	
	With 26 tasks in the dictionary (4 source-only tasks), our approach leads to training $26$ fully supervised task-specific networks, $22 \times 25$ transfer networks in 1$^{st}$ order, and $22 \times {25\choose k}$ for $k^{th}$ order, from which we sample according to the procedure in Sec.~\ref{sec:method}. The total number of transfer functions trained for the taxonomy was $\sim$3,000 which took 47,886 GPU hours on the cloud.
	
	Out of 26 tasks, we use the following 4 as source-only tasks (described in Sec.~\ref{sec:method}) in the experiments: colorization, jigsaw puzzle, in-painting, random projection. However, the method is applicable to an arbitrary partitioning of the dictionary into $\mathcal{T}$ and $\mathcal{S}$. The interactive \href{http://taskonomy.stanford.edu/api}{solver website} allows the user to specify any desired partition.

	\begin{table}
		\centering
		\footnotesize
		\setlength\tabcolsep{1.5pt} 
		\begin{tabular}[]{|l|l l| l | l l | l | l l | }
			\hline
			Task & $avg$ & $rand$ & Task & $avg$ & $rand$ & Task & $avg$ & $rand$ \\
			\hline
			\hline
			Denoising & 100 & 99.9 & Layout & 99.6 & 89.1 & Scene Class. & 97.0 & 93.4 \\ 
			Autoenc. & 100 & 99.8    & 2D Edges & 100 & 99.9 &   Occ. Edges & 100 & 95.4 \\
			Reshading & 94.9 & 95.2 & Pose (fix) & 76.3 & 79.5  &  Pose (nonfix) & 60.2 & 61.9  \\
			Inpainting & 99.9 & -  & 2D Segm. & 97.7 & 95.7 & 2.5D Segm. & 94.2 & 89.4 \\
			Curvature & 78.7 & 93.4  &  Matching & 86.8 & 84.6 & Egomotion & 67.5 & 72.3 \\
			Normals & 99.4 & 99.5 & Vanishing & 99.5 & 96.4   & 2D Keypnt. & 99.8 & 99.4 \\
			Z-Depth & 92.3 & 91.1 &  Distance & 92.4 & 92.1 & 3D Keypnt. & 96.0 & 96.9 \\
			\hline
			Mean & 92.4 & 90.9 & \multicolumn{6}{|c|}{} \\ \hline
		\end{tabular}
		\smallskip
		\caption{\footnotesize{\textbf{Task-Specific Networks' Sanity:} Win rates vs. \emph{random} (Gaussian) network representation readout and statistically informed guess $avg$.}}
		\label{table:full_supervision}
	\end{table}
	
	\textbf{Network Architectures:} We preserved the architectural and training details across tasks as homogeneously as possible to avoid injecting any bias.
	The \textbf{encoder} architecture is identical across all task-specific networks and is a fully convolutional ResNet-50 without pooling.
	All \textbf{transfer} functions include identical shallow networks with 2 conv layers (concatenated channel-wise if higher-order). The loss ($L_t$) and \textbf{decoder}'s architecture, though, have to depend on the task as the output structures of different tasks vary; for all pixel-to-pixel tasks, e.g. normal estimation, the decoder is a 15-layer fully convolutional network; for low dimensional tasks, e.g. vanishing points, it consists of 2-3 FC layers. All networks are trained using the same hyperparameters regardless of task and on exactly the same input images. Tasks with more than one input, e.g. relative camera pose, share weights between the encoder towers. Transfer networks are all trained using the same hyperparameters as the task-specific networks, except that we anneal the learning rate earlier since they train much faster. Detailed definitions of architectures, training process, and experiments with different encoders can be found in the \suppref{supplementary material}.

	\textbf{Data Splits:} Our dataset includes 4 million images. We made publicly available the models trained on full dataset, but for the experiments reported in the main paper, we used a subset of the dataset as the extracted structure stabilized and did not change when using more data (explained in Sec.~\ref{sec:universality}). The used subset is partitioned into training (120k), validation (16k), and test (17k) images, each from non-overlapping sets of buildings. Our task-specific networks are trained on the training set and the transfer networks are trained on a subset of validation set, ranging from 1k images to 16k, in order to model the transfer patterns under different data regimes. In the main paper, we report all results under the 16k transfer supervision regime ($\sim$10\% of the split) and defer the additional sizes to the \suppref{supplementary material} and \href{http://taskonomy.stanford.edu/}{website} (see Sec.~\ref{sec:universality}). Transfer functions are evaluated on the test set. 
	
	\textbf{How good are the trained task-specific networks?} \emph{Win rate (\%)} is the proportion of test set images for which a baseline is beaten. Table \ref{table:full_supervision} provides win rates of the task-specifc networks vs. two baselines. 
	Visual outputs for a random test sample are in Fig.~\ref{fig:outputs}. The high win rates in Table \ref{table:full_supervision} and qualitative results show the networks are well trained and stable and can be relied upon for modeling the task space. See results of applying the networks on a YouTube video frame-by-frame \href{http://taskonomy.stanford.edu/\#models}{here}. A live demo for user uploaded queries is available~\href{http://taskonomy.stanford.edu/tasks}{here}. 
	
	To get a sense of the quality of our networks vs. state-of-the-art task-specific methods, we compared our depth estimator vs. released models of ~\cite{laina2016deeper} which led to outperforming~\cite{laina2016deeper} with a win rate of 88\% and losses of 0.35 vs. 0.47 (further details in the \suppref{supplementary material}). In general, we found the task-specific networks to perform on par or better than state-of-the-art for many of the tasks, though we do not formally benchmark or claim this.

	\begin{figure}
		\includegraphics[width=0.97\columnwidth]{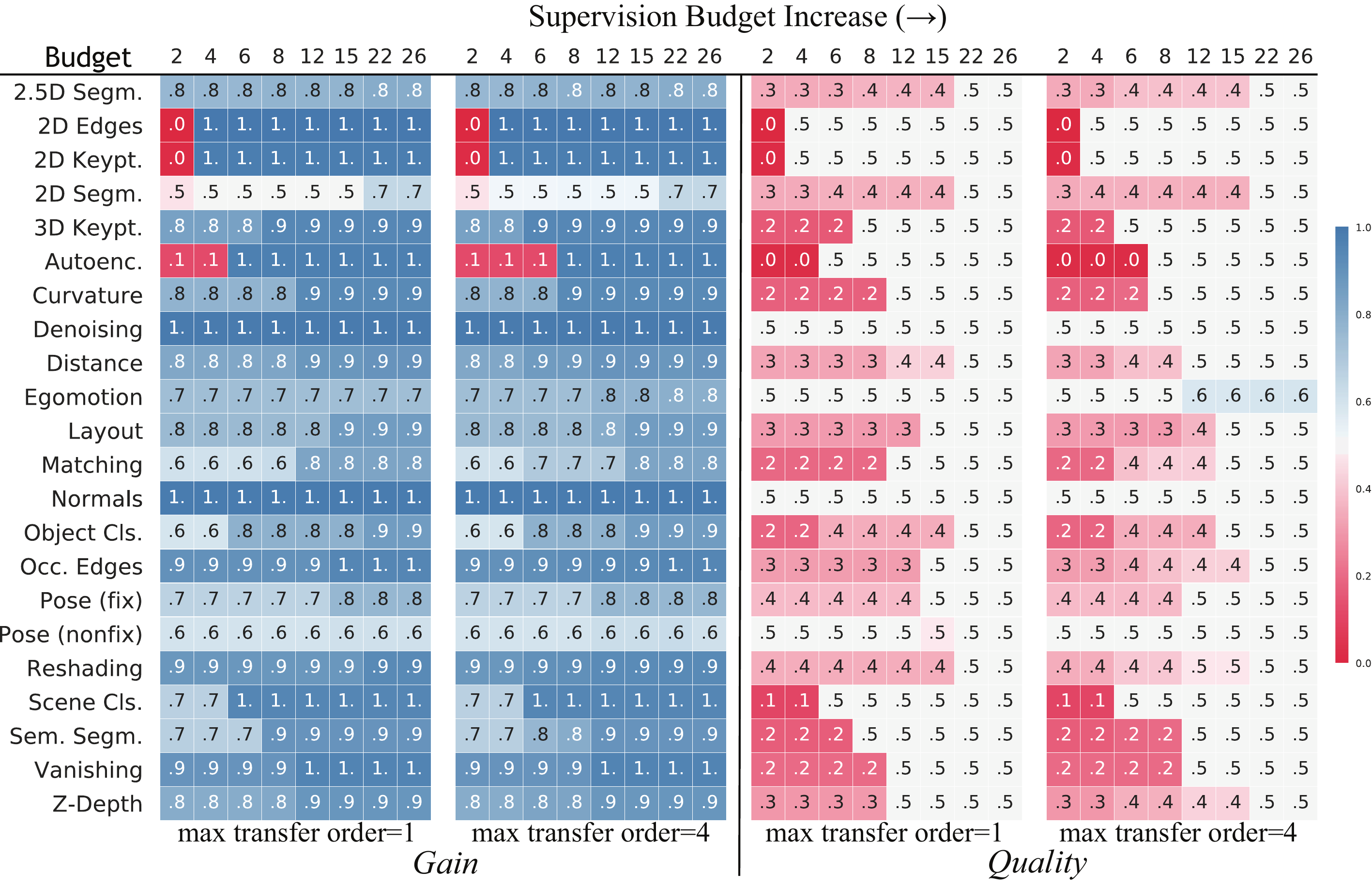}
		\vspace{-9pt}
		\caption{\footnotesize{\textbf{Evaluation of taxonomy computed for solving the full task dictionary}.  Gain (left) and Quality (right) values for each task using the policy suggested by the computed taxonomy, as the supervision budget increases($\rightarrow$). Shown for transfer orders 1 and 4.}}
		\label{fig:quality_and_increase}
	\end{figure}
	
	\subsection{Evaluation of Computed Taxonomies} \label{section:computed_taxonomies}
	Fig.~\ref{fig:taxonomies_over_time} shows the computed taxonomies optimized to solve the full dictionary, i.e. all tasks are placed in $\mathcal{T}$ and $\mathcal{S}$ (except for 4 source-only tasks that are in $\mathcal{S}$ only). This was done for various supervision budgets (columns) and maximum allowed order (rows) constraints. Still seeing transfers to some targets when the budget is 26 (full dictionary) means certain transfers became better than their fully supervised task-specific counterpart.
	
	While Fig.~\ref{fig:taxonomies_over_time} shows the structure and connectivity, Fig.~\ref{fig:quality_and_increase} quantifies the results of taxonomy recommended transfer policies by two metrics of \emph{Gain} and \emph{Quality}, defined as:
	
	\noindent\textbf{Gain:} win rate (\%) against a network trained from scratch using the same training data as transfer networks'. That is, the best that could be done if transfer learning was not utilized. This quantifies the \emph{gained} value by transferring. \\
	\textbf{Quality:} win rate (\%) against a fully supervised network trained with 120k images (gold standard). 
	
	Red~(0) and Blue~(1) represent outperforming the reference method on none and all of test set images, respectively (so the transition Red$\rightarrow$White$\rightarrow$Blue is desirable. White~(.5) represents equal performance to reference).
	
	Each column in Fig.~\ref{fig:quality_and_increase} shows a supervision budget. As apparent, good results can be achieved even when the supervision budget is notably smaller than the number of solved tasks, and as the budget increases, results improve (expected). Results are shown for 2 maximum allowed orders. 

	\definecolor{seaborn}{RGB}{149,177,201}
	\definecolor{seabornbluetext}{RGB}{0,0,0}
	
	\definecolor{seabornred}{RGB}{242,197,201}
	\definecolor{seabornredtext}{RGB}{0,0,0}

	\begin{figure}
		\hspace{-0.6cm}
		\begin{minipage}[b]{0.28\columnwidth}
			\centering
			\includegraphics[height=2.4\columnwidth]{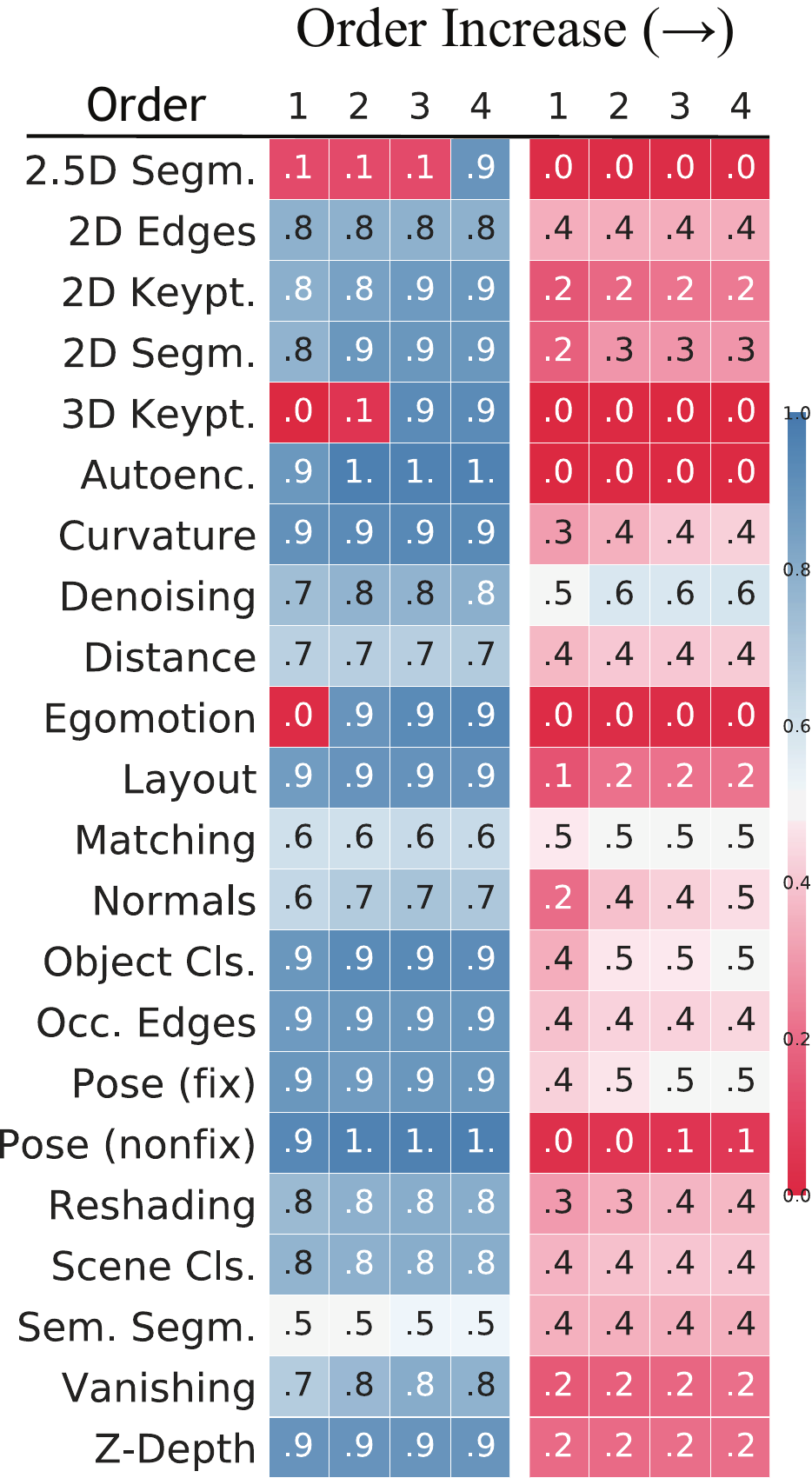}
		\end{minipage}
		\begin{minipage}[b]{0.77\columnwidth}
			\hspace{0.7cm}
			\setlength\tabcolsep{1.5pt} 
			\footnotesize
			\renewcommand{\arraystretch}{0.7}
			\begin{tabular}[b]{ |c|c|c|c|c|c|c|c|c|c| }
				\hline
				Task & \rotatebox[origin=l]{90}{\scriptsize{scratch}} & \rotatebox[origin=l]{90}{\tiny{ImageNet\cite{krizhevsky2012imagenet}}}  & \rotatebox[origin=l]{90}{\tiny{Wang.\cite{wang2015unsupervised}}} & \rotatebox[origin=l]{90}{\tiny{Agrawal.\cite{agrawal2015learning}}} & \rotatebox[origin=l]{90}{\tiny{Zamir.\cite{amir2016generic}}} & \rotatebox[origin=l]{90}{\tiny{Zhang.\cite{zhang2016colorful}}} & \rotatebox[origin=l]{90}{\tiny{Noroozi.\cite{NorooziF16}}} & 
				\rotatebox[origin=l]{90}{\scriptsize{full sup.}} & 
				\rotatebox[origin=l]{90}{\scriptsize{Taxonomy}}\\ \hline
				\multirow{2}{*}{Depth} 	  & \cellcolor{seaborn} \textcolor{seabornbluetext}{88} & \cellcolor{seaborn} \textcolor{seabornbluetext}{88} & \cellcolor{seaborn} \textcolor{seabornbluetext}{93} &\cellcolor{seaborn} \textcolor{seabornbluetext}{89} &\cellcolor{seaborn} \textcolor{seabornbluetext}{88} &\cellcolor{seaborn} \textcolor{seabornbluetext}{84} &\cellcolor{seaborn} \textcolor{seabornbluetext}{86} &\cellcolor{seabornred} \textcolor{seabornredtext}{43} & - \\ \cline{2-10}
				&\tiny{.03} 	 &  \tiny{.04}  &  \tiny{.04}  & \tiny{.03} &\tiny{.04} &\tiny{.03} &\tiny{.03} &\textbf{\tiny{.02}} &\tiny{.02} \\ \hline
				\multirow{2}{*}{Scene Cls.} 	  & \cellcolor{seaborn} \textcolor{seabornbluetext}{80} & \cellcolor{seaborn} \textcolor{seabornbluetext}{52} & \cellcolor{seaborn} \textcolor{seabornbluetext}{83} &\cellcolor{seaborn} \textcolor{seabornbluetext}{74} &\cellcolor{seaborn} \textcolor{seabornbluetext}{74} &\cellcolor{seaborn} \textcolor{seabornbluetext}{71} &\cellcolor{seaborn} \textcolor{seabornbluetext}{75} &\cellcolor{seabornred} \textcolor{seabornredtext}{15} & - \\ \cline{2-10}
				&\tiny{3.30} 	 &  \tiny{2.76}  &  \tiny{3.56}  & \tiny{3.15} &\tiny{3.17} &\tiny{3.09} &\tiny{3.19} &\textbf{\tiny{2.23}} &\tiny{2.63} \\ \hline
				\multirow{2}{*}{Sem. Segm.} 	  & \cellcolor{seaborn} \textcolor{seabornbluetext}{78} & \cellcolor{seaborn} \textcolor{seabornbluetext}{79} & \cellcolor{seaborn} \textcolor{seabornbluetext}{82} &\cellcolor{seaborn} \textcolor{seabornbluetext}{85} &\cellcolor{seaborn} \textcolor{seabornbluetext}{76} &\cellcolor{seaborn} \textcolor{seabornbluetext}{78} &\cellcolor{seaborn} \textcolor{seabornbluetext}{84} &\cellcolor{seabornred} \textcolor{seabornredtext}{21} & - \\ \cline{2-10}
				&\tiny{1.74} 	 &  \tiny{1.88}  &  \tiny{1.92}  & \tiny{1.80} &\tiny{1.85} &\tiny{1.74} &\tiny{1.71} &\textbf{\tiny{1.42}} &\tiny{1.53} \\ \hline
				\multirow{2}{*}{Object Cls.} 	  & \cellcolor{seaborn} \textcolor{seabornbluetext}{79} & \cellcolor{seaborn} \textcolor{seabornbluetext}{54} & \cellcolor{seaborn} \textcolor{seabornbluetext}{82} &\cellcolor{seaborn} \textcolor{seabornbluetext}{76} &\cellcolor{seaborn} \textcolor{seabornbluetext}{75} &\cellcolor{seaborn} \textcolor{seabornbluetext}{76} &\cellcolor{seaborn} \textcolor{seabornbluetext}{76} &\cellcolor{seabornred} \textcolor{seabornredtext}{34} & - \\ \cline{2-10}
				&\tiny{4.08} 	 &  \tiny{3.57}  &  \tiny{4.27}  & \tiny{3.99} &\tiny{3.98} &\tiny{4.00} &\tiny{3.97} &\textbf{\tiny{3.26}} &\tiny{3.46} \\ \hline
				\multirow{2}{*}{Normals} 	  & \cellcolor{seaborn} \textcolor{seabornbluetext}{97} & \cellcolor{seaborn} \textcolor{seabornbluetext}{98} & \cellcolor{seaborn} \textcolor{seabornbluetext}{98} &\cellcolor{seaborn} \textcolor{seabornbluetext}{98} &\cellcolor{seaborn} \textcolor{seabornbluetext}{98} &\cellcolor{seaborn} \textcolor{seabornbluetext}{97} &\cellcolor{seaborn} \textcolor{seabornbluetext}{97} &\cellcolor{seabornred}6 & - \\ \cline{2-10}
				&\tiny{.22} 	 &  \tiny{.30}  &  \tiny{.34}  & \tiny{.28} &\tiny{.28} &\tiny{.23} &\tiny{.24} &\textbf{\tiny{.12}} &\tiny{.15} \\ \hline
				\multirow{2}{*}{2.5D Segm.} 	  & \cellcolor{seaborn} \textcolor{seabornbluetext}{80} & \cellcolor{seaborn} \textcolor{seabornbluetext}{93} & \cellcolor{seaborn} \textcolor{seabornbluetext}{92} &\cellcolor{seaborn} \textcolor{seabornbluetext}{89} &\cellcolor{seaborn} \textcolor{seabornbluetext}{90} &\cellcolor{seaborn} \textcolor{seabornbluetext}{84} &\cellcolor{seaborn} \textcolor{seabornbluetext}{87} &\cellcolor{seabornred} \textcolor{seabornredtext}{40} & - \\ \cline{2-10}
				&\tiny{.21} 	 &  \tiny{.34}  &  \tiny{.34}  & \tiny{.26} &\tiny{.29} &\tiny{.22} &\tiny{.24} &\textbf{\tiny{.16}} &\tiny{.17} \\ \hline
				\multirow{2}{*}{Occ. Edges} 	  & \cellcolor{seaborn} \textcolor{seabornbluetext}{93} & \cellcolor{seaborn} \textcolor{seabornbluetext}{96} & \cellcolor{seaborn} \textcolor{seabornbluetext}{95} &\cellcolor{seaborn} \textcolor{seabornbluetext}{93} &\cellcolor{seaborn} \textcolor{seabornbluetext}{94} &\cellcolor{seaborn} \textcolor{seabornbluetext}{93} &\cellcolor{seaborn} \textcolor{seabornbluetext}{94} &\cellcolor{seabornred} \textcolor{seabornredtext}{42} & - \\ \cline{2-10}
				&\tiny{.16} 	 &  \tiny{.19}  &  \tiny{.18}  & \tiny{.17} &\tiny{.18} &\tiny{.16} &\tiny{.17} &\textbf{\tiny{.12}} &\tiny{.13} \\ \hline
				\multirow{2}{*}{Curvature} 	  & \cellcolor{seaborn} \textcolor{seabornbluetext}{88} & \cellcolor{seaborn} \textcolor{seabornbluetext}{94} & \cellcolor{seaborn} \textcolor{seabornbluetext}{89} &\cellcolor{seaborn} \textcolor{seabornbluetext}{85} &\cellcolor{seaborn} \textcolor{seabornbluetext}{88} &\cellcolor{seaborn} \textcolor{seabornbluetext}{92} &\cellcolor{seaborn} \textcolor{seabornbluetext}{88} &\cellcolor{seabornred} \textcolor{seabornredtext}{29} & - \\ \cline{2-10}
				&\tiny{.25} 	 &  \tiny{.28}  &  \tiny{.26}  & \tiny{.25} &\tiny{.26} &\tiny{.26} &\tiny{.25} &\textbf{\tiny{.21}} &\tiny{.22} \\ \hline
				\multirow{2}{*}{Egomotion} 	  & \cellcolor{seaborn} \textcolor{seabornbluetext}{79} & \cellcolor{seaborn} \textcolor{seabornbluetext}{78} & \cellcolor{seaborn} \textcolor{seabornbluetext}{83} &\cellcolor{seaborn} \textcolor{seabornbluetext}{77} &\cellcolor{seaborn} \textcolor{seabornbluetext}{76} &\cellcolor{seaborn} \textcolor{seabornbluetext}{74} &\cellcolor{seaborn} \textcolor{seabornbluetext}{71} &\cellcolor{seaborn} \textcolor{seabornbluetext}{59} & - \\ \cline{2-10}
				&\tiny{8.60} 	 &  \tiny{8.58}  &  \tiny{9.26}  & \tiny{8.41} &\tiny{8.34} &\tiny{8.15} &\tiny{7.94}  &\tiny{7.32} &\textbf{\tiny{6.85}} \\ \hline
				\multirow{2}{*}{Layout} 	  & \cellcolor{seaborn} \textcolor{seabornbluetext}{80} & \cellcolor{seaborn} \textcolor{seabornbluetext}{76} & \cellcolor{seaborn} \textcolor{seabornbluetext}{85} &\cellcolor{seaborn} \textcolor{seabornbluetext}{79} &\cellcolor{seaborn} \textcolor{seabornbluetext}{77} &\cellcolor{seaborn} \textcolor{seabornbluetext}{78} &\cellcolor{seaborn} \textcolor{seabornbluetext}{70} &\cellcolor{seabornred} \textcolor{seabornredtext}{36} & - \\ \cline{2-10}
				&\tiny{.66} 	 &  \tiny{.66}  &  \tiny{.85}  & \tiny{.65} &\tiny{.65} &\tiny{.62} &\tiny{.54} &\textbf{\tiny{.37}} &\tiny{.41} \\ \hline
			\end{tabular}
		\end{minipage}
		\vspace{-0.6cm}					
		\caption{\footnotesize{\textbf{Generalization to Novel Tasks}. Each row shows a novel test task. Left: Gain and Quality values using the devised ``all-for-one" transfer policies for novel tasks for orders 1-4. Right: Win rates (\%) of the transfer policy over various self-supervised methods, ImageNet features, and scratch are shown in the colored rows. Note the large margin of win by taxonomy. The uncolored rows show corresponding loss values.} \vspace{-.2cm}}
		\label{fig:task_specific_quality_gain}
		\vspace{-0cm}
	\end{figure}
	
	\subsection{Generalization to Novel Tasks}
	\label{sec:novel_task}
	The taxonomies in Sec.~\ref{section:computed_taxonomies} were optimized for solving all tasks in the dictionary. In many situations, a practitioner is interested in a single task which even may not be in the dictionary. Here we evaluate how taxonomy transfers to a novel out-of-dictionary task with little data.
	
	This is done in an all-for-one scenario where we put one task in $\mathcal{T}$ and all others in $\mathcal{S}$. The task in $\mathcal{T}$ is target-only and has no task-specific network. Its limited data (16k) is used to train small transfer networks to sources. This basically \emph{localizes} where the target would be in the taxonomy.
	
	
	Fig.~\ref{fig:task_specific_quality_gain} (left) shows the \emph{Gain} and \emph{Quality} of the transfer policy found by the BIP for each task.  
	Fig.~\ref{fig:task_specific_quality_gain} (right) compares the taxonomy suggested policy against some of the best existing self-supervised methods~\cite{wang2015unsupervised,zhang2016colorful,NorooziF16,amir2016generic,agrawal2015learning}, ImageNet FC7 features~\cite{krizhevsky2012imagenet}, training from scratch, and a fully supervised network (gold standard).

	The results in Fig.~\ref{fig:task_specific_quality_gain} (right) are noteworthy. The large win margin for taxonomy shows that carefully selecting transfer policies depending on the target is superior to fixed transfers, such as the ones employed by self-supervised methods. ImageNet features which are the most popular off-the-shelf features in vision are also outperformed by those policies. Additionally, though the taxonomy transfer policies lose to fully supervised networks (gold standard) in most cases, the results often get close with win rates in 40\% range. These observations suggests the space has a rather predicable and strong structure. For graph visualization of the all-for-one taxonomy policies please see the \suppref{supplementary material}. The \href{http://taskonomy.stanford.edu/api}{solver website} allows generating the taxonomy for arbitrary sets of target-only tasks.

	\subsection{Significance Test of the Structure}
	\label{sec:significance}
\begin{figure}
		\includegraphics[width=1\columnwidth]{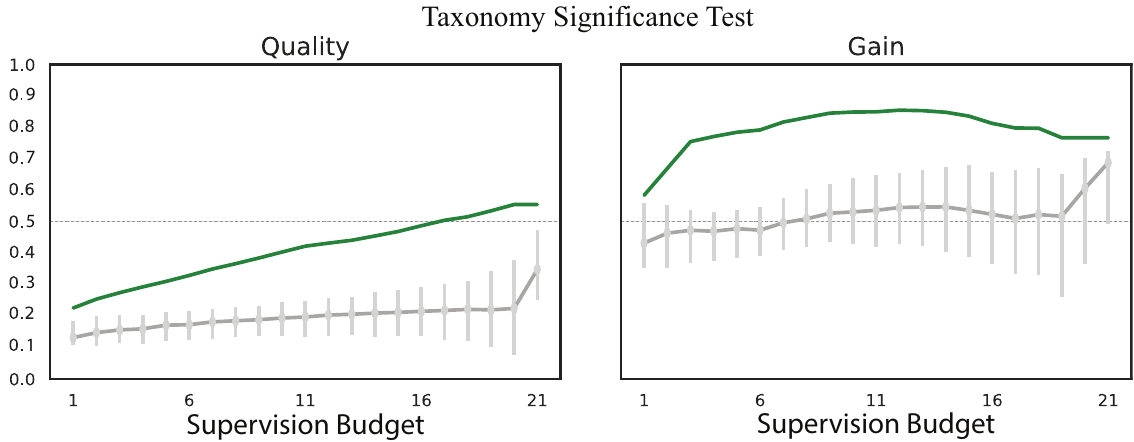}
		\caption{\footnotesize{\textbf{Structure Significance.} Our taxonomy compared with random transfer policies (random feasible taxonomies that use the maximum allowable supervision budget). Y-axis shows \emph{Quality} or \emph{Gain}, and X-axis is the supervision budget. Green and gray represent our taxonomy and random connectivities, respectively. Error bars denote 5$^{th}$--95$^{th}$ percentiles.} }
		\label{fig:taxonomy_significance}
	\end{figure}

	The previous evaluations showed good transfer results in terms of \emph{Quality} and \emph{Gain}, but how crucial is it to use our taxonomy to choose smart transfers over just choosing \emph{any} transfer? In other words, how \emph{significant/strong} is the discovered structure of task space? Fig.~\ref{fig:taxonomy_significance} quantifies this by showing the performance of our taxonomy versus a large set of taxonomies with random connectivities. Our taxonomy outperformed all other connectivities by a large margin signifying both existence of a strong structure in the space as well as a good modeling of it by our approach. Complete experimental details is available in \suppref{supplementary material}.

	\subsection{Evaluation on MIT Places \& ImageNet}
	\label{sec:external_datasets}
\begin{figure}
		\centering
		\begin{minipage}[b]{1\textwidth}
			\includegraphics[width=0.48\columnwidth]{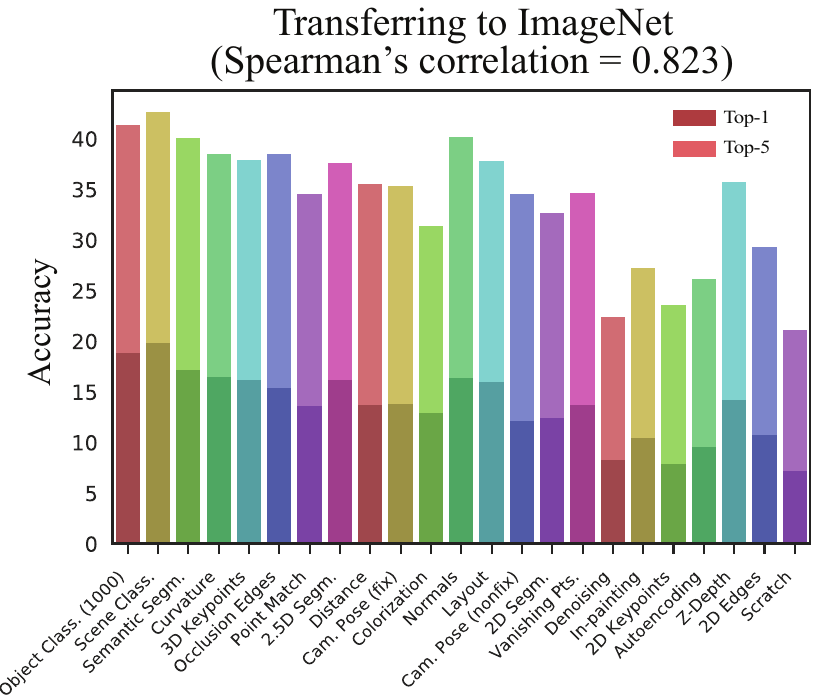}
			\includegraphics[width=0.48\columnwidth]{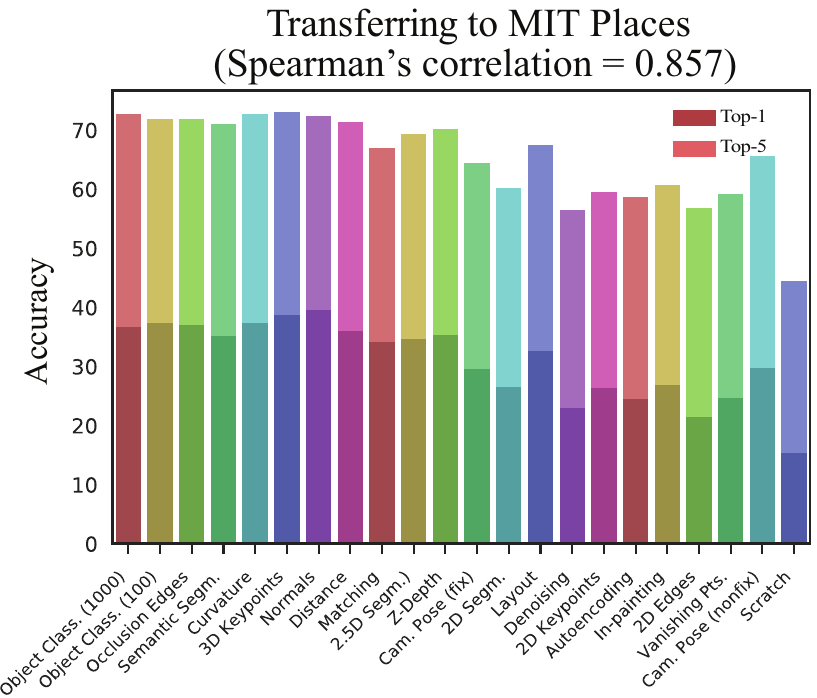}
		\end{minipage}
		\vspace{-0.5cm}
		\caption{ \footnotesize{\textbf{Evaluating the discovered structure on other datasets: ImageNet~\cite{russakovsky2015imagenet} (left) for object classification and MIT Places~\cite{zhou2014learning} (right) for scene classification.} Y-axis shows accuracy on the external benchmark while bars on x-axis are ordered by taxonomy's predicted performance based on our dataset. A monotonically decreasing plot corresponds to preserving identical orders and perfect generalization.}}
		\label{fig:imagenet_transfer}
	\end{figure}

	To what extent are our findings dataset dependent, and would the taxonomy change if done on another dataset? We examined this by finding the ranking of all tasks for transferring to two target tasks of object classification and scene classification on our dataset. We then fine tuned our task-specific networks on other datasets (MIT Places~\cite{zhou2014learning} for scene classification, ImageNet~\cite{russakovsky2015imagenet} for object classification) and evaluated them on their respective test sets and metrics. Fig.~\ref{fig:imagenet_transfer} shows how the results correlate with taxonomy's ranking from our dataset. 
	The Spearman's rho between the taxonomy ranking and the Top-1 ranking is 0.857 on Places and 0.823 on ImageNet showing a notable correlation. See \suppref{supplementary material} for complete experimental details. 
	
	\subsection{Universality of the Structure}
	\label{sec:universality}
	We employed a computational approach with various design choices. It is important to investigate how specific to those the discovered structure is. We did stability tests by computing the variance in our output when making changes in one of the following system choices: \textbf{I. architecture of task-specific networks, II. architecture of transfer function networks, III. amount of data  available for training transfer networks, IV. datasets, V. data splits, VI. choice of dictionary}. Overall, despite injecting large changes (e.g. varying the size of training data of transfer functions by 16x, size and architecture of task-specific networks and transfer networks by 4x), we found the outputs to be remarkably stable leading to almost no change in the output taxonomy computed on top. Detailed results and experimental setup of each tests are reported in the \suppref{supplementary material}. 
	

	\subsection{Task Similarity Tree}
	\label{sec:tree}
	Thus far we showed the task space has a structure, measured this structure, and presented its utility for transfer learning via devising transfer policies. This structure can be presented in other manners as well, e.g. via a metric of similarity across tasks. Figure~\ref{fig:tree} shows a similarity tree for the tasks in our dictionary. This is acquired from agglomerative clustering of the tasks based on their transferring-out behavior, i.e. using columns of normalized affinity matrix $P$ as feature vectors for tasks. The tree shows how tasks would be hierarchically positioned w.r.t. to each other when measured based on providing information for solving other tasks; the closer two tasks, the more similar their role in transferring to other tasks. Notice that the 3D, 2D, low dimensional geometric, and semantic tasks are found to cluster together using a fully computational approach, which matches the intuitive expectations from the structure of task space. The transfer taxonomies devised by BIP are consistent with this tree as BIP picks the sources in a way that all of these modes are quantitatively best covered, subject to the given budget and desired target set.

	
	\definecolor{transfer2d}{HTML}{1f77b4}
	\definecolor{transfer3d}{HTML}{2ca02c}
	\definecolor{transferLowDim}{HTML}{d62728}
	\definecolor{transferSemantic}{HTML}{9467bd}
	
	\begin{figure}
		\includegraphics[width=1\columnwidth]{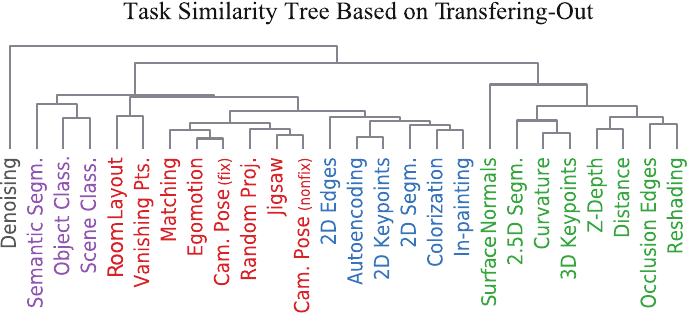}
		\caption{\footnotesize{\textbf{Task Similarity Tree.} Agglomerative clustering of tasks based on their transferring-out patterns (i.e. using columns of normalized affinity matrix as task features). \textcolor{transfer3d}{3D}, \textcolor{transfer2d}{2D}, \textcolor{transferLowDim}{low dimensional geometric}, and \textcolor{transferSemantic}{semantic} tasks clustered together using a fully computational approach.}}
		\label{fig:tree}
	\end{figure}

	\section{Limitations and Discussion}
	We presented a method for modeling the space of visual tasks by way of transfer learning and showed its utility in reducing the need for supervision. The space of tasks is an interesting object of study in its own right and we have only scratched the surface in this regard. We also made a number of assumptions in the framework which should be noted.  
	
	\emph{Model Dependence:} We used a computational approach and adopted neural networks as our function class. Though we validated the stability of the findings w.r.t various architectures and datasets, it should be noted that the results are in principle model and data specific. The current model also does not include a principled mechanism for handling uncertainty or probabilistic reasoning.
	
	\emph{Compositionality:} We performed the modeling via a set of common human-defined visual tasks. It is natural to consider a further compositional approach in which such common tasks are viewed as \emph{observed samples} which are composed of computationally found latent (sub)tasks. 
	
	\emph{Space Regularity:} We performed modeling of a dense space via a sampled dictionary. Though we showed a good tolerance w.r.t. to the choice of dictionary and transferring to out-of-dictionary tasks, this outcome holds upon a proper sampling of the space as a function of its regularity. More formal studies on properties of the computed space is required for this to be provably guaranteed for a general case. 
	
	\emph{Transferring to Non-visual and Robotic Tasks:} Given the structure of the space of visual tasks and demonstrated transferabilities to novel tasks, it is worthwhile to question how this can be employed to develop a perception module for solving downstream tasks which are not entirely visual, e.g. robotic manipulation, but entail solving a set of (a priori unknown) visual tasks.
	
	\emph{Lifelong Learning}: We performed the modeling in one go. In many cases, e.g. lifelong learning, the system is evolving and the number of mastered tasks constantly increase. Such scenarios require augmentation of the structure with expansion mechanisms based on new beliefs.
	\\ \\ 
	\noindent\textbf{Acknowledgement:} We acknowledge the support of~NSF (DMS-1521608), MURI (1186514-1-TBCJE), ONR MURI (N00014-14-1-0671), Toyota(1191689-1-UDAWF), ONR MURI (N00014-13-1-0341), Nvidia, Tencent, a gift by Amazon Web Services, a Google Focused Research Award.


	{\small
		\bibliographystyle{ieee}
		\bibliography{egbib}
	}
	

\clearpage
\makeatletter
\let\maketitle\suppsavedmaketitle
\let\@maketitle\suppsavedatmaketitle
\let\title\suppsavedtitle
\let\author\suppsavedauthor
\let\and\suppsavedand
\let\thanks\suppsavedthanks
\let\date\suppsaveddate
\makeatother
\pagestyle{empty}
\setlength{\textfloatsep}{20pt plus 2pt minus 4pt}
\setlist[enumerate]{itemsep=0pt}
\setcounter{page}{1}
\makeatletter
\let\suppOrigEveryPageAnchor\Hy@EveryPageAnchor
\def\Hy@EveryPageAnchor{\begingroup\let\suppOrigAnchor\hyper@@anchor\def\hyper@@anchor##1{\suppOrigAnchor{supp.##1}}\suppOrigEveryPageAnchor\endgroup}
\long\def\backrefxxx#1#2#3{\hyperlink{supp.page.#1}{#1}}
\makeatother
\setcounter{footnote}{0}
\setcounter{section}{0}
\setcounter{figure}{0}
\setcounter{table}{0}
\setcounter{equation}{0}
\renewcommand{\theHsection}{supp.\arabic{section}}
\renewcommand{\theHfigure}{supp.\arabic{figure}}
\renewcommand{\theHtable}{supp.\arabic{table}}
\renewcommand{\theHequation}{supp.\arabic{equation}}

\title{Taskonomy: Disentangling Task Transfer Learning \\ \large{\underline{Supplementary Material}}}
	
	\author{Amir R. Zamir$^{1,2}$ \;\; Sasha Sax$^{1}$\thanks{equal.} \;\; William Shen$^{1}$\footnotemark[1] \;\;  Leonidas Guibas$^{1}$ \;\; Jitendra Malik$^{2}$ \;\; Silvio Savarese$^{1}$\vspace{10pt}\\ 
		$^1$ Stanford University  \;\;  
		$^2$ University of California, Berkeley\vspace{10pt}\\ 
		\textcolor{blue}{\url{http://taskonomy.stanford.edu/}\vspace{0pt}}
	}
	
	\maketitle
	\hypertarget{supplement.start}{}

\begin{abstract}
The following items are provided in the supplementary material:
 \renewcommand{\labelenumii}{\Roman{enumii}}
\begin{enumerate}[topsep=0pt,itemsep=-1ex,partopsep=1ex,parsep=1ex]
    \item Higher-order transfer experiments.
    \item A tiny-data taxonomy.
    \item Optimal transfer policies for selected tasks.
    \item Our networks as feature extractors on MIT Places.
    \item BIP Solver as API.
    \item Significance testing of taxonomy.
    \item Qualitative results of network output.
    \item Agglomerative clustering of tasks.
    \item Analysis of transitive transfers.
    \item Experimental setup details.
    \item Task dictionary definition and discussion.
\end{enumerate}
\end{abstract}

\section{Higher-Order Transfers}

In the main paper, low-order transfers (orders 2-4) often achieved superior performance to first-order transfers. What happens as we continue to increase the order? Do we get the best performance from using the entire bank of task-specific networks?

In addition to training many first-order and low-order (orders 2-4) transfers, we trained one transfer function for each order 5-25. Each high-order transfer function was trained on the combined representations of the top $k$ first-order transfers. The high-order transfers are available in the provided API. The calculated distances matrix including higher-order transfers is shown in figure \ref{fig:higher_order_affinities}.

\begin{figure}
 \includegraphics[width=\columnwidth]{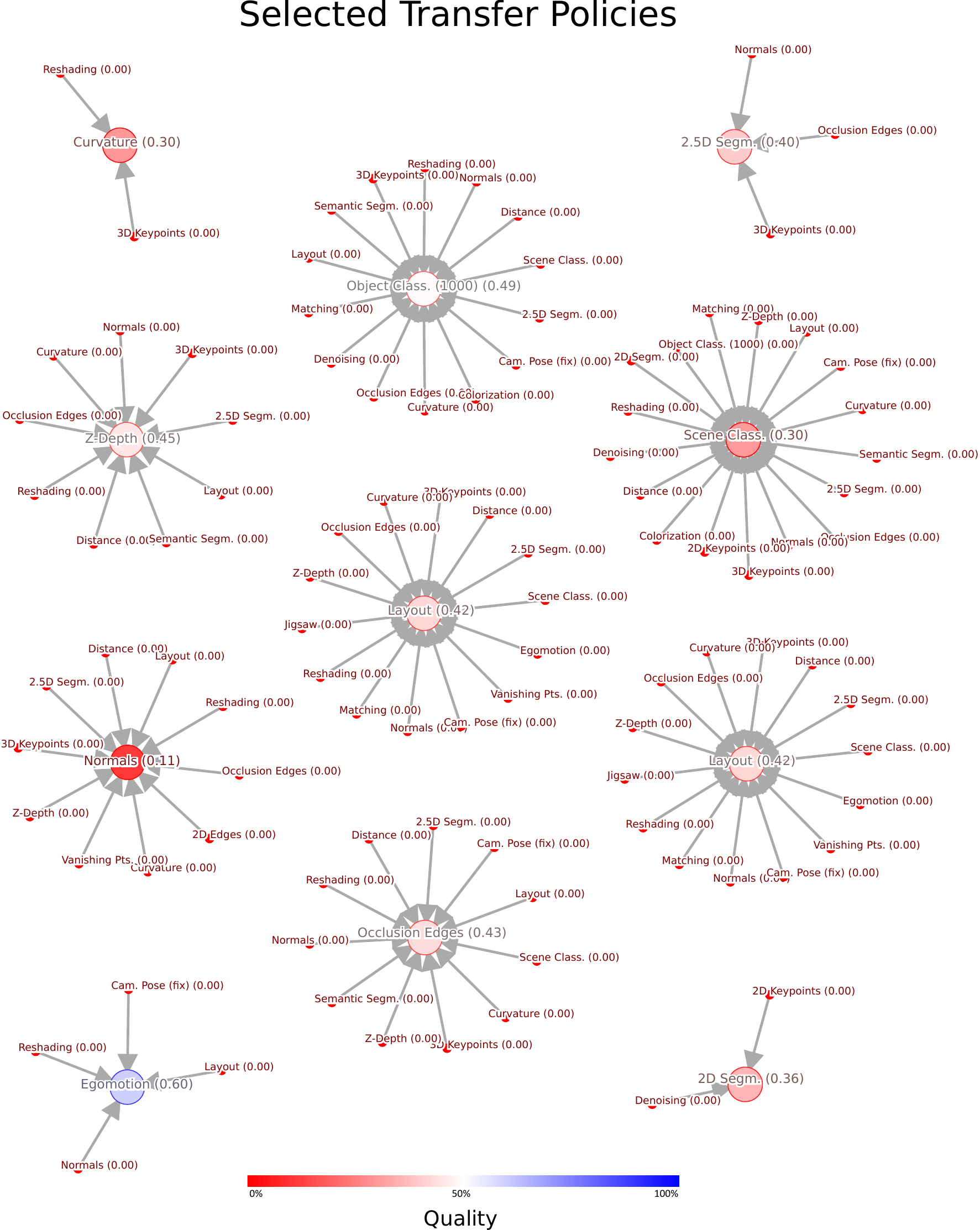}
    \caption{\footnotesize{\textbf{Selected policies.} For tasks in the table form Figure 10 of the main paper, we show the best policies for transferring to that task. The center node of each cluster is the target task, and the surrounding nodes at the tail of each arrow are the suggested source tasks.}}
  \label{fig:esp-task-policy}
\end{figure}

\begin{figure*}
 \includegraphics[width=\textwidth]{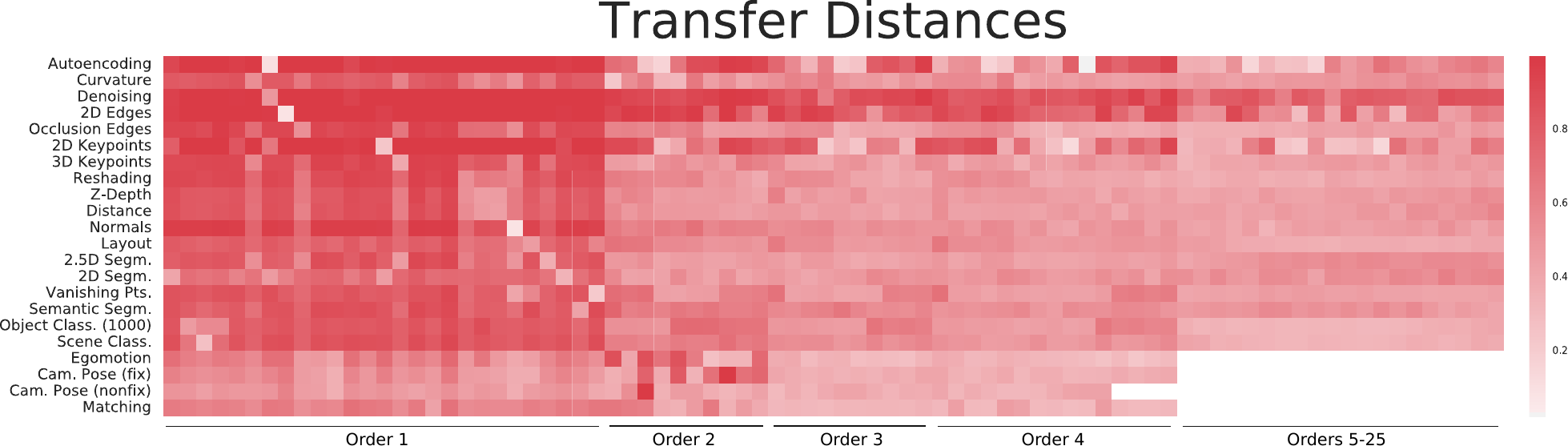}
    \caption{\footnotesize{\textbf{Higher Order Distances.} Calculated task distances for all 16k transfers. The columns are sorted by transfer order and the orders are shown along the x-axis. Each row is a target task.} }
  \label{fig:higher_order_affinities}
\end{figure*}

\section{Tiny Data Taxonomy} 
\begin{figure}
 \includegraphics[width=\columnwidth]{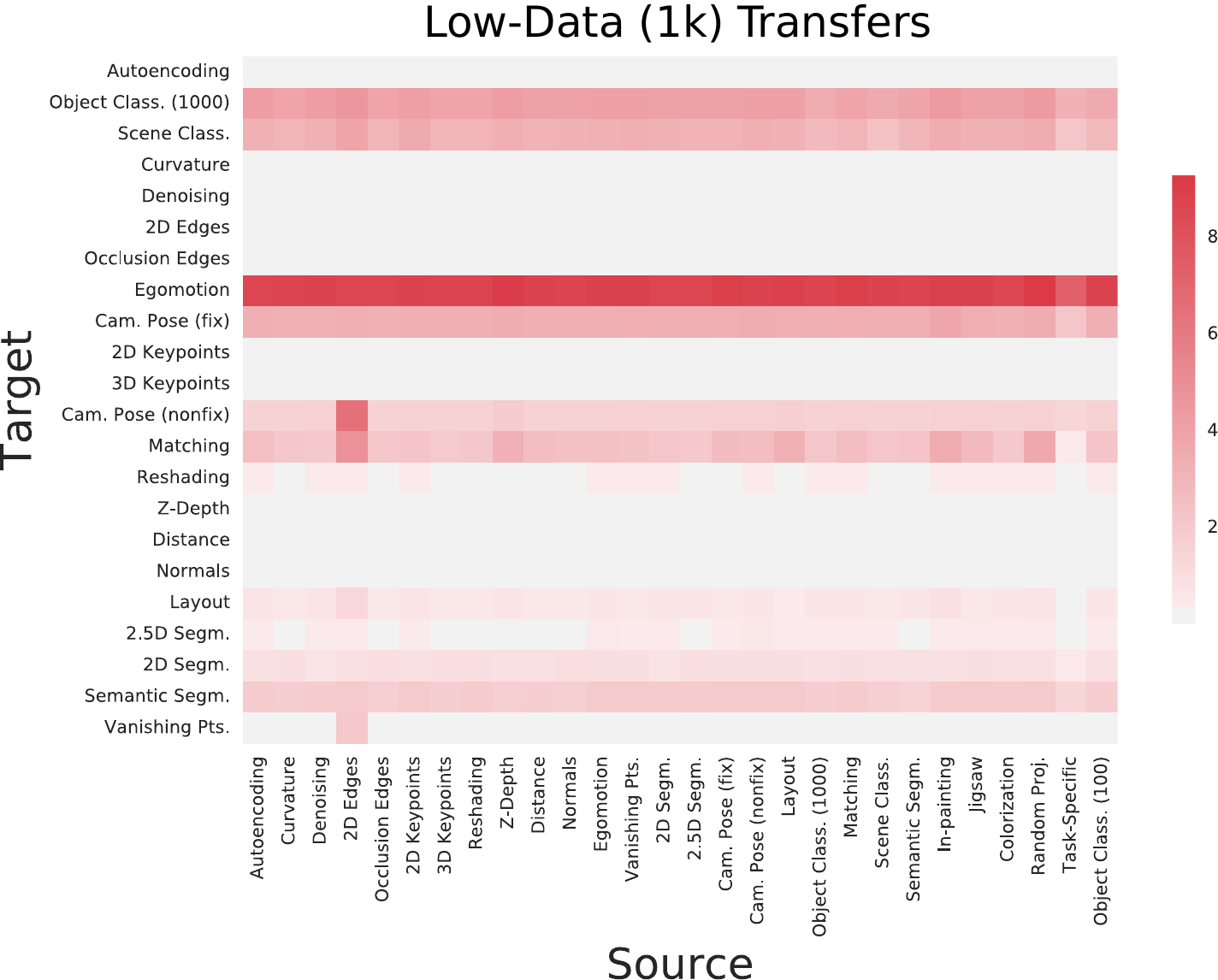}
    \caption{\footnotesize{\textbf{Low data transfers.} Expected loss after training transfer functions with 1k samples (shown before AHP normalization). This is a low-data version of the transfer loss matrix in the main paper.} }
  \label{fig:low_data_mat}
\end{figure}

As the amount of supervision of the transfer networks increases, the performance margin between them decreases. Eventually, full supervision is possible. Clearly, the amount of data available for training transfers is an important factor.

We present a low-data taxonomy experiment in which the transfer networks were trained on 1k data points, as opposed to the main experiments which used 16k data points. The distance matrix for these transfers is in figure \ref{fig:low_data_mat}, and the transfers themselves are accessible in the provided API.

\section{Special Task Policies}

In the main paper, we showed the best transfer performance of our networks against some commonly used baselines. In figure \ref{fig:esp-task-policy}, we now show the collections of transfers that beat the baselines shown in figure 9 in the main paper. In some cases, higher-order transfers outperformed the ones in figure 9, and we show those instead. 

\section{Feature Performance on MIT Places}

\begin{figure}
 \includegraphics[width=\columnwidth]{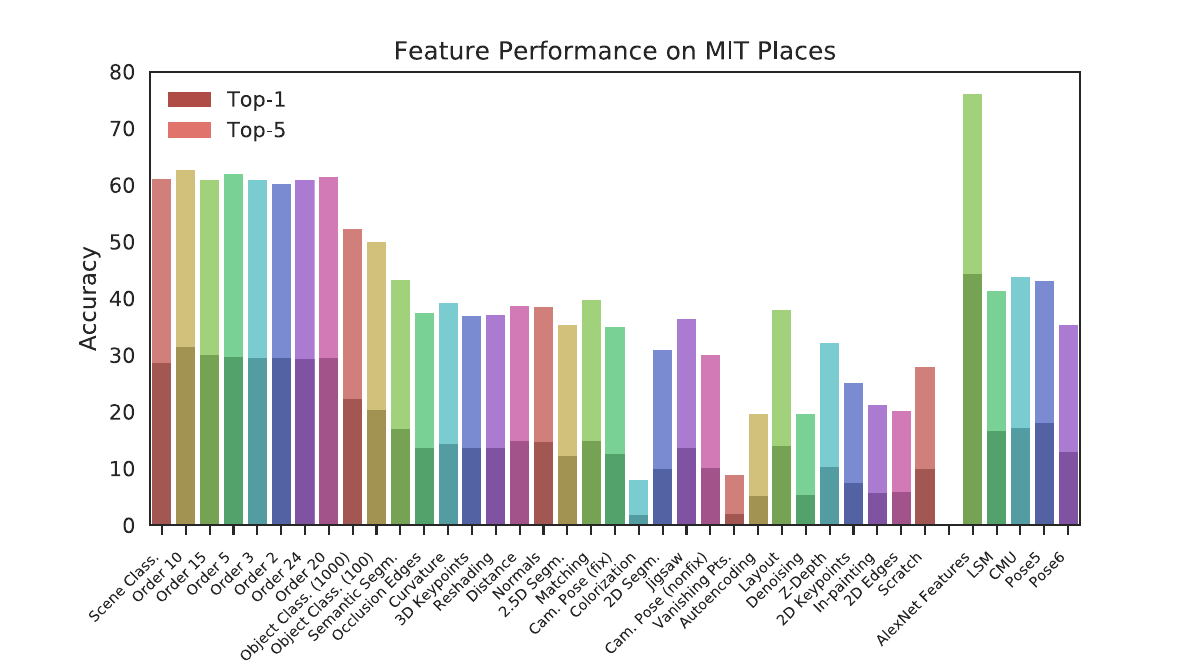}
\caption{\footnotesize{\textbf{Feature extraction.} Performance of readout functions trained on MIT Places. We compare our networks to common baselines. The x-axis is ordered according to transfer performance on our dataset, and the y-axis shows Top-1 and Top-5 accuracy on MIT Places. The sources on the left were trained on our dataset, while the right portion contains external baselines.} }
  \label{fig:mit_feature_extraction}
\end{figure}

When data labels are scarce, AlexNet and other pretrained networks are commonly used as feature extractors that provide input for a small readout function. We compare the performance of our task-specific features to AlexNet features on 16k images from the 63 selected classes on the MIT Places dataset. While AlexNet was trained on 1M hand-labeled images, our task-specific networks were trained on 120k images where labels were generated automatically. The final results of the MIT feature extraction experiment are shown in figure \ref{fig:mit_feature_extraction}.

\section{BIP Solver as API}

The BIP solver allows users to query our results and quickly compute the best transfer taxonomy for their particular use case. In this sense it can acts as an API, and we refer to it as the Transfer Learning API, or just the API. \href{http://taskonomy.stanford.edu/api}{http://taskonomy.stanford.edu/api} provides access to this API via a webpage and allows interactively entering the arguments and computing the corresponding transfer taxonomy. Qualitative transfer results corresponding to the computed taxonomy exemplified over a YouTube video is also shown.



\section{Sensitivity to Choice of Dictionary}

\begin{figure}
	
  \includegraphics[width=1\columnwidth]{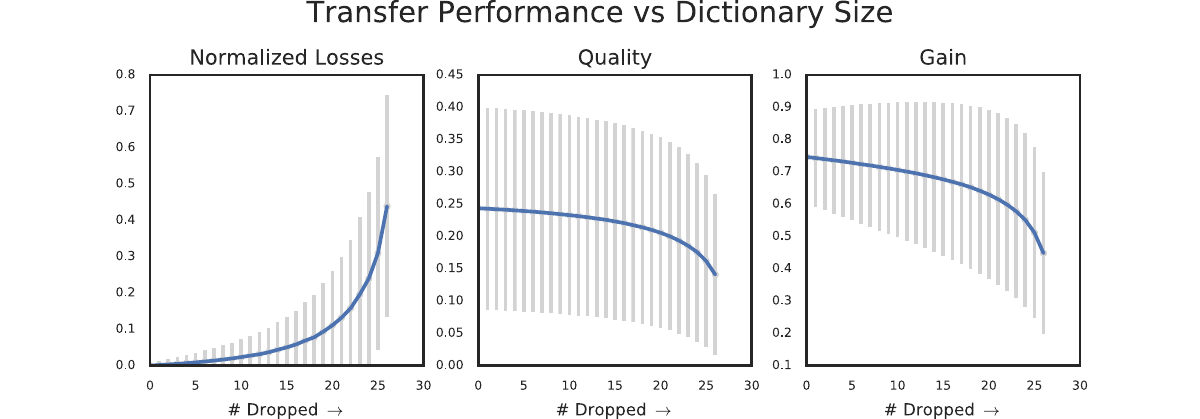}
  \caption{\footnotesize{\textbf{Drop-k performance.} Expected performance of transferring to a task after dropping $k$ tasks from our dictionary. The grey bars show the standard deviation of the performance.} }
  \label{fig:drop_k_transfer}
\end{figure}

How universal are our findings with respect to the modeling dictionary? Is our coverage of the task space sufficiently dense? We evaluate this by randomly dropping some tasks from the dictionary and evaluating first-order transfer performance to tasks, and averaging over tasks. If our dictionary is dense, we should see a gradual drop in performance until our dictionary becomes too sparse, at which point performance will degrade rapidly. Results are shown in figure \ref{fig:drop_k_transfer}.

Additionally, in order to give a more comprehensive idea of the opportunity costs of choosing sub-optimal transfers, we show the losses, quality, and gain of all first-order transfers in figure \ref{fig:overlay}.

\section{Transfer via Transitivity}

Perhaps transfer can be done in a transitive manner: first learn task $A$ from images, and then $B$ from $A$, and finally $C$ from $B$. Intuitively, the performance $A \rightarrow C$ might improve if existing neural network machinery can take advantage of the additional computational power and supervision in the transitive  $A \rightarrow B \rightarrow C$.

\begin{figure}
  \includegraphics[width=1.0\columnwidth]{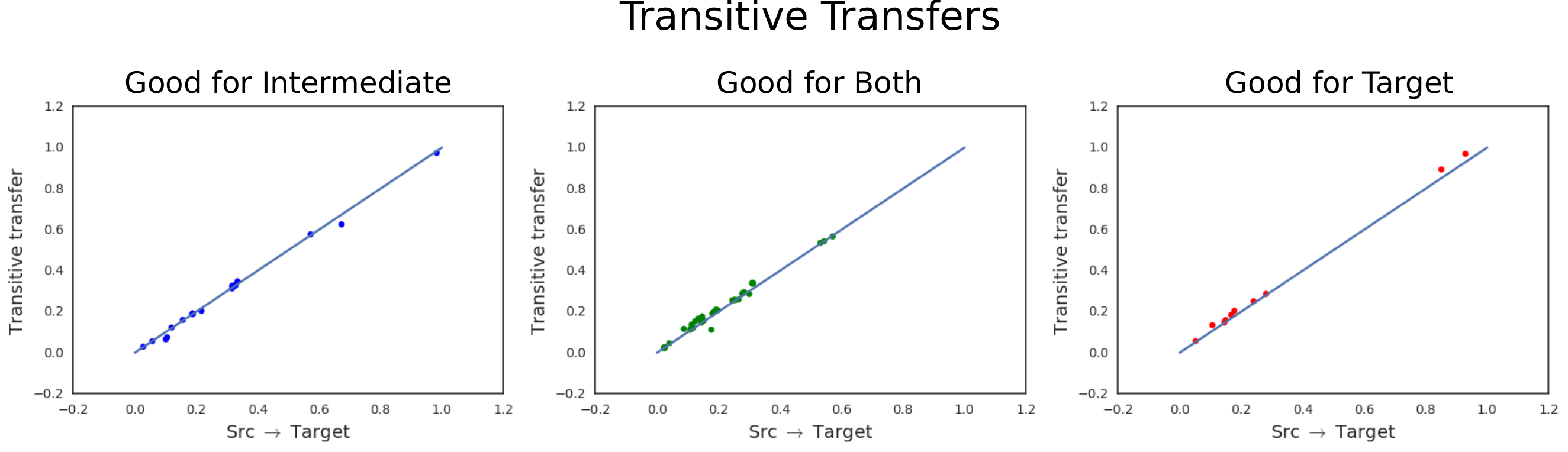}
  \caption{\footnotesize{\textbf{Transitivity.} Loss after transitive transfer is on the y-axis and loss after direct ($A \rightarrow C$) transfer is on the x-axis. The line x=y is shown. Red dots are transfers which transferred well to $C$ but not to $B$ in first-order. Blue dots transferred well to the target, $C$, but not to $B$. Green dots transferred well to both.} }
  \label{fig:transitivity}
\end{figure}

See figure \ref{fig:transitivity} for the a comparison of the two transfer types. Based on the intermediate transfer $B$, we show three cases for transitive transfers: $B$ transferred well to only $A$, only $C$, or both $A$ and $C$. We found that the loss for transfers $A \rightarrow B \rightarrow C$ is very close to the loss for $A \rightarrow C$. This means that the largest contributor to performance on $C$ is the representations learned for $A$, and that neither a doubling of the transfer computation nor the additional supervision for a proxy task has a strong effect on the performance. Since these results suggest that NNs are unable to effectively use transitive transfers, even if we modeled them, we are enforcing the taxonomy to be free of chains longer than length 1.

\section{I. Model Specificness.}
In principle, the findings are model dependent as a natural consequence of employing a fully computational formulation. However, we performed a set of experiments with varying transfer network and task-specific network architectures to quantify model independence within the function class of neural nets. The AHP affinity matrices turned out to be very similar (left) and the variance matrices (right) show pretty low values, signifying a high invariance.

\begin{figure}[H]
\hspace*{-0.9pt}\includegraphics{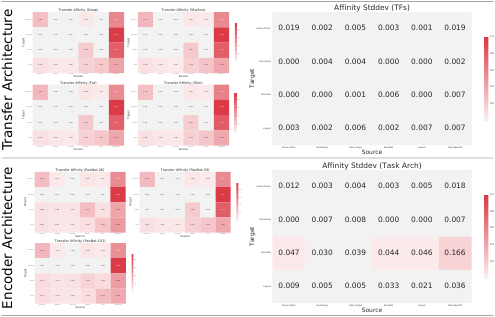}
\caption{\footnotesize{\textbf{Model Specificity.} (left) Affinity matrices. (right) Standard Deviation of Affinities.} }
\label{fig:model_specificity}
\end{figure}

\section{Comparison to per-task state-of-the-art methods}
We reported the win rate of our task specific models vs. two statistical baselines of average estimator \textit{avg} (the best statistically informed guess) and a supervised readout network trained on a random (Gaussian) network's representation (\textit{rand}) in the main paper. To get a sense of the quality of our networks vs. state-of-the-art task-specific methods, we compared our depth estimator vs. released models of Fully Convolutional Residual Network \cite{supp:laina2016deeper}. The results are provided in the table below showing our task specific network outperformed \cite{supp:laina2016deeper}. In general, we found the task-specific networks to perform on par or better than state-of-the-art for many of the tasks, though we do not formally benchmark or claim this.

\begin{center}
\begin{tabular}{|c|c|c|}
\hline
Evaluation Pair & Win Rate & Loss \\
\hline\hline
{\footnotesize Ours (120k training images) v FCRN} & 80\% & \textbf{0.39} v 0.47 \\
{\footnotesize Ours (3M training images) v FCRN} & 88\% & \textbf{0.35} v 0.47 \\
\hline
\end{tabular}\vspace{5.75pt}
\end{center}




\section{Evaluation on MIT Places \& ImageNet: Experimental Details}

Does our ranking of transfers hold across datasets? Is the taxonomy ordering something innate to the tasks, or an artifact of our data? We evaluated this on MIT Places and ImageNet, and found a very strong correlation between our ordering and the one on the new dataset; the results are reported in the main paper. The setup for these experiments is as follows:

\textbf{Transfer to MIT Places}:
We evaluated the order on Places-365 in the following way. For each task in our dataset, we reinitialized a new transfer function, attached it to the task-specific encoder, and trained the network normally on 16k images from the 63 classes in the Home+Hotel and Work subgroups from Places-365, classes which were likely to be present in our dataset.

Note that the places training regimen is different from the one used to form the taxonomy, and the training and evaluation occur on different datasets. Despite this, correlation between the two orders was very high.

\textbf{Transfer to ImageNet}:
Similar to our Places experiment, we selected a subset of categories likely to appear in our dataset (e.g. appliances, furniture). This had little effect on the ordering on our dataset. Then we distilled a trained ResNet-152~\cite{supp:he2016deep} and train a transfer function to predict the activations.

\section{Baselines}
To evaluate the performance of our networks, we compare our performance against 6 common baselines. Four of the baselines are trained on different external datasets: 
 \renewcommand{\labelenumii}{\Roman{enumii}}
\begin{enumerate}
\item \textbf{ImageNet fc7\cite{supp:krizhevsky2012imagenet}}, with representation size 4096, is trained on ImageNet~\cite{supp:ILSVRC15} 1000l way classification.
\item \textbf{Wang et al.~\cite{supp:wang2015unsupervised}}, with representation size 1024, is trained on 100K videos from YouTube for unsupervised patch triplet ranking.
\item \textbf{Agrawal et al.~\cite{supp:agrawal2015learning}}, with representation shape $13 \times 13 \times 256$, is an unsupervised method trained on KITTI to do egomotion in  outdoor environments.
\item \textbf{Zamir et al.~\cite{supp:amir2016generic}}, with representation size 500, is trained to do both relative camera pose prediction and point matching.
\end{enumerate}

Two of the baselines are trained as standard tasks on our dataset: Colorization~\cite{supp:Zhang2016} and Context Prediction (Jigsaw)~\cite{supp:NorooziF16}, and details are defined in sec. \ref{sec:task_dict}.

In the comparison to baseline tasks shown in Figure 10 of the main paper, the experimental setup was as follows: we extracted features for each baseline network, and then trained transfer functions to predict the labels of other tasks using the extracted representations. In the case where the baseline features were not of size 2048, we added an extra fully-connected layer. We then evaluated the quality and gain of our networks over these baselines using our standard ordinal metric. 

We also compare our networks to these baselines on other datasets for which our networks were not trained. We compare the features extracted by our networks to those extracted by the baselines on MIT Places \cite{supp:MITPlaces} by training readout functions. Results are shown in \ref{fig:mit_feature_extraction}. 

\section{Network Architecture}

We used Tensorflow for our implementation.

\textbf{Encoder:} For all tasks we modified a ResNet-50 encoder with no average-pooling and replace the last stride 2 convolution with stride 1. This gives us an output shape of $16 \times 16 \times 2048$; we use a $3\times 3$ convolution to transform the output to our final representation, which is of shape $16 \times 16 \times 8$. 

\textbf{Decoder:} The shape of the decoder depends on the task. For pixels-to-pixels prediction, we used a 15-layer fully convolutional decoder: 5 convolutional layers, and then alternating convolution and convolution\_transpose layers. For low-dimensional tasks such as Vanishing Points, our decoder was two or three fully-connected layers with dropout applied after all but the last one. For all layers (except output), we used batchnorm.  

\textbf{Transfer Function:} Our transfer functions have a tower structure, and multiple towers have their output concatenated channelwise. Our towers are feedforward with the following set of functions: first clip extreme values to 5-sigma and then do batch renormalization, followed by two convolutional layers. In the case where the readout function takes an image as input, these layers are stride-4 dilated convolutions. As our transfer functions may use both an image and a representation, we used two towers in every transfer function -- if there was no input image, we fed a representation to both towers. 

\textbf{Loss Functions:} For pixels-to-pixels prediction tasks we used a CGAN in addition to the standard loss function. The GAN training began after 25000 steps, and the discriminator was 5 layers with $10 \times$ the standard weight decay. The loss was a linear combination of the $0.996 \times$ the standard loss and $0.004 \times$ the GAN loss. We found that including even this small amount of GAN loss helped to sharpen the images. 

\textbf{Hyperparameters:} We used ADAM with a learning rate of $10^{-4}$ and annealed the learning rate by a factor of 10 after 80000 steps. We used batch size 32 and weight decay $2 \times 10^{-6}$.

\textbf{Data:} Out of more than 1M images collected, we used 120k from 87 buildings to train task-specific networks. The transfer networks were trained on 16k images from a separate 24 buildings. The taxonomy was generated from these transfer networks on a third set of 17k images from 13 buildings.

\section{Pseudo-semantics Annotations}
\label{sec:pseudo}
As we do not have semantic annotations on our dataset, we gathered pseudo-semantic annotations by using knowledge distillation~\cite{supp:hinton2015distilling} approach. That is, we labeled our dataset using the output of state-of-the-art large fully-supervised network (ResNet-151) for semantic objects (using ImageNet, 100 applicable classes), scene categories (using MIT Places~\cite{supp:MITPlaces}, 63 indoor workplace and home classes in MIT Places scene hierarchy), and semantic object segmentation (using COCO~\cite{supp:MSCoco}, 17 applicable classes). In a user study, human annotators indicated that only 6.3\% of images have no plausible label in their top-5 classes in Scene categories, showing that such labels are reliable enough and effective for being a gateway to modeling semantics in task space. The quality of the semantic labels can be seen in the provided sample of qualitative results or in the video. The list of selected classes for COCO and ImageNet are available {\href{https://github.com/StanfordVL/taskonomy/tree/master/taskbank/assets/web_assets/pseudosemantics}{here}}.
 
\section{Detailed Definition of Task Dictionary}\label{sec:task_dict}

Task definitions are presented in alphabetical order:

\textbf{Autoencoding}
PCA is a widely used method of understanding data by finding a low-dimensional latent representation. Autoencoding \cite{supp:Hinton504} is a nonlinear generalization of PCA that was originally proposed with transfer learning in mind: better downstream performance through autoencoding pretraining.

\textbf{Colorization}~\cite{supp:Zhang2016}
Colorization requires taking a gray-scale image and predicting the original colors. It is an unsupervised task, but also one that is semantic-aware\cite{supp:Chia:2011:SCI:2070781.2024190}. For example, predicting the color of a fruit is simple once the fruit is identified.

\textbf{Context Encoding}
Context Encoding was first introduced by Pathak et al. \cite{supp:pathakCVPR16context} and is a version of autoencoding where a large portion of the input is masked from the model. In order to fill in the occluded area, the model must reason about the scene geometry and semantics. Similar to colorization, it is an unsupervised-yet-still-semantic task.  


\textbf{Content Prediction (Jigsaw)}\cite{supp:NorooziF16}
A discriminative version of Context Encoding, Jigsaw \cite{supp:NorooziF16} requires a network to unscramble a permuted tiling of the input image.

\textbf{Curvature Estimation}
Curvature-based features are excellent for identification because they are invariant under rigid transformations. Curvature is known to be important in visual processing--so much so that the Macaque visual cortex has a dedicated curvature processing region \cite{supp:Yue19082014}.

\textbf{Denoising}
It is desirable for similar inputs to have similar representations, but representations learned by autoencoding are excessively sensitive to perturbations in the input. Denoising \cite{supp:Vincent:2008:ECR:1390156.1390294} (autoencoding) encourages limited invariance by mapping slightly perturbed inputs to the unperturbed input.

\textbf{Depth Estimation, Euclidean}
Depth estimation is an important task, useful for detecting proximity to obstacles and items of interest. It is also a useful intermediate step for agents to localize themselves in 3D space. Euclidean depth is measured as the distance from each pixel to the camera's optical center. 

\textbf{Depth Estimation, Z-Buffer}
As opposed to Euclidean depth estimation, researchers typically use z-buffer depth which is defined as the distance to the camera plane. This is not the way that humans typically perceive depth, but is included because this is the standard formulation and all of our depth-derived tasks are derived from \textit{z-buffer}. 


\textbf{Edge Detection (2D)}
Edge detection is historically a fundamental task in computer vision. Edges are commonly used as an intermediate representation or as a feature in a larger processing pipeline. We include the output of a Canny edge detector without nonmax suppression (to make the task learnable by neural networks).

\textbf{Edge Detection (3D)}
As opposed to 2D edges, we define 3D edges as ``occlusion edges,'' or edges where an object in the foreground obscures something behind it. 2D edges respond to changes in texture, but 3D edges are features which depend only on the 3D geometry and are invariant to color and lighting. 

\textbf{Keypoint Detection (2D)}
Keypoint detection has a long history in computer vision, and is useful for many, many tasks. Keypoint algorithmns usually consist of two parts, both a keypoint detector and some local patch descriptor which is invariant across multiple images \cite{supp:sift,supp:surf,supp:orb}. 2D keypoint detection encourages the network to identify locally important regions of an image, and \textit{point matching} encourages the network to learn feature descriptors. Identifying keypoints is frequently still a first step in a larger visual pipeline. We use the output of SURF \cite{supp:surf} (before nonmax suppression) as our ground-truth.

\textbf{Keypoint Detection (3D)}
3D keypoints are similar to 2D keypoints except that they are derived from 3D data and therefore account for scene geometry. They are often invariant to informative (but possibly distracting) features such as textures \cite{supp:iss,supp:narf,supp:Mian2010,supp:Zaharescu,supp:Knopp2010}. We use the output of the NARF \cite{supp:narf} algorithm (before nonmax suppression) as our 3D keypoint ground-truth.

\textbf{Point Matching}
Deep networks trained for point matching learn feature descriptors that prove useful for downstream tasks. Point matching has applications in fine-grained classification \cite{supp:yao2012} and object recognition \cite{supp:lowe}, multi-view reconstruction \cite{supp:Seitz2006} and structure from motion \cite{supp:Molton:etal:BMVC2004}, wide baseline matching \cite{supp:WangNY09}, SLAM \cite{supp:Se02mobilerobot} and visual odometry\cite{supp:zhu2007}. 


\textbf{Relative Camera Pose Estimation, Non-Fixated}
The famous ``Kitten Carousel" experiment by Held and Hein~\cite{supp:HeldAndHeinKittens} suggested that taking action is crucial for strong perception. Although more recent works call the original conclusion into question \cite{supp:human_infants_egomotion}, the ability to localize oneself remains important for locomotion. For two different views with the same optical centers, we try to predict the 6-DOF relative camera pose (yaw, pitch, roll, xyz translation) between them.

\textbf{Relative Camera Pose Estimation, Fixated}
We also include a simpler variant of camera pose estimation for which the center pixel of the two inputs is always the same physical 3D point. This problem is simpler in the sense that there are only five degrees of freedom. 

\textbf{Relative Camera Pose Estimation, Triplets (Egomotion)}
Videos are a common object of study in computer vision (e.g. visual odometry \cite{supp:Geiger2012CVPR,supp:nisterOdometry}) and they provide dense data with high redundancy. We therefore include camera pose matching for input triplets with a fixed center point. With three images, models have a greater ability to match points for accurate localization. 

\textbf{Reshading}
One way to infer scene geometry is ``shape from shading'' \cite{supp:ShapeFromShadingBarronAndMalik} using the intrinsic image decomposition $I = A \cdot S$, where S is a shading function parameterized by lighting and depth. This decomposition is thought to be useful in human visual perception \cite{supp:ShadingPerception1996}. We define reshading as follows: Given an RGB image, the label is the shading function $S$ that results from having a single point light at the camera origin, and $S$ is multiplied by a constant fixed albedo. 

\textbf{Room Layout Estimation}
Estimating and aligning a 3D bounding box is a mid-level task that includes vanishing point estimation as a sub-problem, and has applications for robotic navigation \cite{supp:Silberman2012}, scene reconstruction \cite{supp:IzadiniaSS16}, and augmented reality \cite{supp:DeToneMR16}. A variant of room layout estimation was used in the LSUN room layout challenge \cite{supp:yu15lsun}, but that formulation is ill-posed when there is camera roll or when no room corners are in view. Instead, we offer a formulation that remains well-defined regardless of the camera pose and field of view. This task includes some semantic information such as `what constitutes a room' while simultaneously also including scene geometry.


\textbf{Segmentation, Unsupervised (2D))}
Gestalt psychologists proposed principles of grouping as a mechanism through which humans learn to perceive the world as a set of coherent objects \cite{supp:Wertheimer1923-WERLOO}. Normalized cuts \cite{supp:Shi:2000:NCI:351581.351611} are one method for segmenting images into perceptually similar groups, and we include this Gestalt task in our dictionary. 

\textbf{Segmentation, Unsupervised (2.5D))}
Segmentation 2.5D uses the same algorithm as 2D, but the labels are computed jointly from the RGB image, the aligned depth image, and the aligned surface normals image. Therefore the 2.5D segmentation applies the principles not just to the world as it seems (in the RGB image), but also to the world as it is (ground-truth 3D). 2.5D segmentation incorporates information about the scene geometry that is not directly present in the RGB image but that is readily inferred by humans.

\textbf{Surface Normal Estimation}
Surface normal estimation is thought to be crucial for spatial cognition. For example, objects can only be placed on surfaces with upwards-facing normals. Even for locomotion, a point with horizontal-facing normals indicates that it cannot be easily traversed. Surface normals are computed directly from the 3D mesh. 

\textbf{Vanishing Point Estimation}
An consequence of perspective, vanishing points offer useful information about the scene geometry \cite{supp:MiksikRoadVPs,supp:KongRoadVPs} and are well studied. Vanishing points prove particularly useful in a Manhattan world \cite{supp:ManhattanWorld,supp:Zhang2016,supp:BazinVPs} where there are three dominant vanishing points corresponding to an X, Y, and Z axis. This assumption is usually met in urban environments. For each model we analytically find these three vanishing points and include them as labels. 

\textbf{Semantic Learning through Knowledge distillation}
While our dataset does not include semantic annotations, semantic understanding comprises a large and important component of modern computer vision. Therefore, we add pseudo-semantic annotations through knowledge distillation \cite{supp:hinton2015distilling}. We distill knowledge from state-of-the-art models \cite{supp:He2015,supp:li2016fully} trained on ImageNet \cite{supp:ILSVRC15} and MS-COCO \cite{supp:LinMBHPRDZ14} by using them to annotate our dataset and then supervising models with those annotations. See section \ref{sec:pseudo} for the details of knowledge distillation process.

\begin{enumerate}
	
	\item \textbf{Classification, Semantic (1000-classes)}
	Semantic object recognition is a fundamental component of visual perception. Children learn at an early age to classify objects after seeing just a handful of examples \cite{supp:addie}. For semantic classification, we distill knowledge form a pretrained ResNet-152 \cite{supp:He2015} (trained on ImageNet) by supervising our model with the ResNet activations. 
	
	\item \textbf{Classification Semantic (100-classes)}
	Many of the classes in ImageNet never appear in our dataset (e.g. animals, sports). We therefore manually select classes which appear in our dataset -- and this happened to be 100 classes. For 100-way classification, we distill knowledge from these 100 activations only.

	\item \textbf{Segmentation, Semantic}
	Models and agents need to know more than just what they are looking at -- they also need to be able to locate the object. Therefore, we include knowledge distillation from the semantic segmentation model in \cite{supp:li2016fully}, trained on MS COCO.~\cite{supp:TaskonomyTaskTransfer2017}
\end{enumerate}

\begin{figure*}
\includegraphics[width=\textwidth]{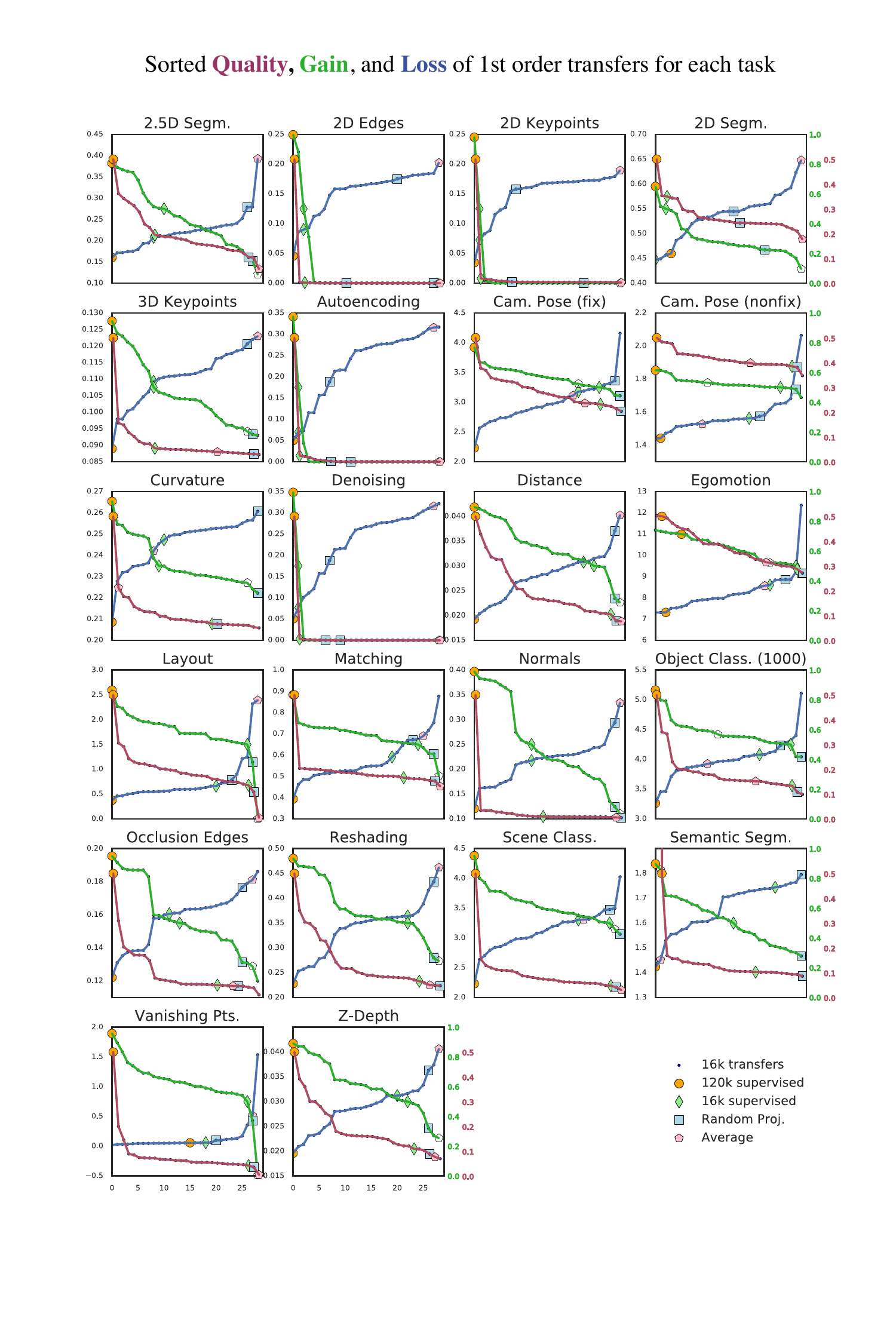}
\refstepcounter{figure}
\label{fig:overlay}
\end{figure*}


\end{document}